\pdfoutput=1

\documentclass[11pt]{article}

\usepackage[]{naacl2021}

\usepackage{times}
\usepackage{latexsym}

\usepackage[T1]{fontenc}

\usepackage[utf8]{inputenc}

\usepackage{microtype}

\usepackage{tikz-qtree}
\usepackage{adjustbox}
\usepackage{tikz}
\usetikzlibrary{intersections}
\usepackage{tkz-euclide}

\usepackage{CJKutf8}

\definecolor{lightslategray}{rgb}{0.47, 0.53, 0.6}
\definecolor{skyblue}{RGB}{70, 130, 180}
\definecolor{fontgray}{RGB}{44, 62, 80}
\definecolor{myred}{RGB}{235, 47, 6} 
\definecolor{myblue}{RGB}{0, 168, 255}
\definecolor{lavendergray}{rgb}{0.77, 0.76, 0.82}
\definecolor{myyellow}{RGB}{235, 237, 239}
\definecolor{lightorange}{RGB}{244, 122, 96}
\definecolor{myback}{RGB}{198, 220, 147}
\definecolor{myg}{RGB}{196, 132, 48}
\definecolor{myx}{RGB}{128, 140, 158}
\definecolor{lowgreen}{RGB}{186, 220, 88} 
\definecolor{littledarkblue}{RGB}{48, 51, 107}
\definecolor{mygray}{RGB}{158,158,158}
\definecolor{fruitpurple}{RGB}{108, 92, 231}
\definecolor{river}{RGB}{52,152,219}
\definecolor{purple}{RGB}{142, 68, 173} 

\usepackage{amsmath}
\usepackage{multirow}
\usepackage{booktabs}
\usepackage{tikz-dependency}
\usepackage{pgfplots}
\usepackage{dashrule}
\usepackage{booktabs,arydshln}
\usepackage{scalerel}
\makeatletter
\def\adl@drawiv#1#2#3{%
        \hskip.5\tabcolsep
        \xleaders#3{#2.5\@tempdimb #1{1}#2.5\@tempdimb}%
                #2\z@ plus1fil minus1fil\relax
        \hskip.5\tabcolsep}
\newcommand{\cdashlinelr}[1]{%
  \noalign{\vskip\aboverulesep
           \global\let\@dashdrawstore\adl@draw
           \global\let\adl@draw\adl@drawiv}
  \cdashline{#1}
  \noalign{\global\let\adl@draw\@dashdrawstore
           \vskip\belowrulesep}}
\makeatother

\def\pnode [#1]#2{
	\node[regular polygon,regular polygon sides=4, minimum size=1pt,fill=gray,#1, inner sep = 1.2pt] (#2) {};
} 
\tikzset{middlefactor/.style={decoration={
			markings,
			mark= at position #1 with {\pnode[]{}} 
		},postaction={decorate}},
	middlefactor/.default=0.5
}

\newcommand{\qed}{\nobreak \ifvmode \relax \else
	\ifdim\lastskip<1.5em \hskip-\lastskip
	\hskip1.5em plus0em minus0.5em \fi \nobreak
	\vrule height0.65em width0.5em depth0.25em\fi}

\usetikzlibrary{arrows,decorations.pathmorphing,backgrounds,positioning,fit,petri,shapes.misc, arrows.meta,shapes.geometric,decorations.markings,calc,shadows.blur,decorations.pathreplacing,quotes,matrix,shapes.symbols,patterns, decorations.pathreplacing, calc}

\pgfdeclarelayer{background}
\pgfdeclarelayer{foreground}
\pgfsetlayers{background,foreground,depgroups,main}
\newcommand{\squishlist}{
	\begin{list}{$\bullet$}
		{ \setlength{\itemsep}{0pt}
			\setlength{\parsep}{3pt}
			\setlength{\topsep}{3pt}
			\setlength{\partopsep}{0pt}
			\setlength{\leftmargin}{1.5em}
			\setlength{\labelwidth}{1em}
			\setlength{\labelsep}{0.5em} } }

	\newcounter{Lcount}
	\newcommand{\squishlisttwo}{
		\begin{list}{\arabic{Lcount}. }
			{ \usecounter{Lcount}
				\setlength{\itemsep}{0pt}
				\setlength{\parsep}{0pt}
				\setlength{\topsep}{0pt}
				\setlength{\partopsep}{0pt}
				\setlength{\leftmargin}{2em}
				\setlength{\labelwidth}{1.5em}
				\setlength{\labelsep}{0.5em} } }
		
		\newcommand{\squishend}{
	\end{list} }

\usepackage[normalem]{ulem}

\title{Fusing Structured Information into LSTM for Named Entity Recognition}

\title{Integrating Structured Information into Contextual Information for Named Entity Recognition}

\title{Learning to Integrate Different Features for Named Entity Recognition}

\title{Integrating Different Types of Features for Named Entity Recognition}

\title{Better Feature Combination for Named Entity Recognition}

\title{Better Feature Integration for Named Entity Recognition}

\newcommand*{\affaddr}[1]{#1} 
\newcommand*{\affmark}[1][*]{\textsuperscript{#1}}

\author{%
Lu Xu\affmark[1, 2]\thanks{$*$  Lu Xu is under the Joint PhD Program between Alibaba and Singapore University of Technology and Design. The work was done when Zhanming Jie was a PhD student in Singapore University of Technology and Design.},
Zhanming Jie\affmark[1, 3], Wei Lu\affmark[1], Lidong Bing\affmark[2]\\
\affaddr{\affmark[1] StatNLP Research Group, Singapore University of Technology and Design}\\
\affaddr{\affmark[2] DAMO Academy, Alibaba Group}~~ \affaddr{\affmark[3] ByteDance}\\
\tt{xu\_lu@mymail.sutd.edu.sg, allan@bytedance.com}\\
\tt{ luwei@sutd.edu.sg, l.bing@alibaba-inc.com}\\
}
\renewcommand\footnotemark{}

\begin{document}

\maketitle
\begin{abstract}

It {\color{black}has been} shown that named entity recognition (NER) could benefit from incorporating the long-distance structured information captured by dependency trees.
We believe this is because both types of features -- the contextual information captured by the linear sequences and the structured information captured by the dependency trees may complement each other.
However, existing approaches largely focused on stacking the LSTM and graph neural networks such as graph convolutional networks (GCNs) for building improved NER models,
where the exact interaction mechanism between the two {\color{black}different} types of features is not very clear, and the performance gain does not appear to be significant.
In this work, we propose a \textcolor{black}{simple and robust} solution to incorporate both types of features with our Synergized-LSTM (Syn-LSTM), which clearly captures how the two types of features interact.
We conduct extensive experiments on several standard datasets across four languages.
The results demonstrate that the proposed model achieves better performance than previous approaches while requiring fewer parameters.
Our further analysis demonstrates that our model can capture longer dependencies compared with  strong baselines.\footnote{We make our code publicly available at \url{https://github.com/xuuuluuu/SynLSTM-for-NER}.}
\end{abstract}

\section{Introduction}
Named entity recognition (NER) is one of the most fundamental and important tasks in natural language processing (NLP). 
While  the literature~\cite{peters2018deep,akbik2018coling,devlin2019bert} largely focuses on training deep language models to improve the contextualized word representations, previous studies show that the structured information such as interactions between {\color{black}non-adjacent} words can also be important for NER~\cite{finkel-etal-2005-incorporating,jie2017efficient,aguilar2019dependency}. 

\begin{figure}[t!]
	\centering
	\adjustbox{max width=1.0\linewidth}{
		\begin{tikzpicture}[node distance=1.0mm and 1.0mm, >=Stealth, 
		wordnode/.style={draw=none, minimum height=5mm, inner sep=0pt},
		chainLine/.style={line width=0.8pt,-, color=fontgray},
		entbox/.style={draw=none, rounded corners, line width=1pt, fill=river!40},
		entboxc/.style={draw=none, rounded corners, line width=1pt}
		]
		
		\matrix (sent) [matrix of nodes, nodes in empty cells, execute at empty cell=\node{\strut};]
		{
			\textit{Precision} &[-1mm] \textit{Castparts} &[-1mm] \textit{Corp.} & \textit{\textbf{,}} &[-1mm] \textit{Portlan} & \textit{\textbf{,}} &[-1mm] \textit{will} &[-1mm] \textit{begin} &[-1mm] \textit{trading} &[-1mm] \textit{with} &[-1mm] \textit{the} &[-1mm] \textit{symbol} & \textit{PCP} & \textbf{\textit{.} }\\
		};

		\draw [chainLine, ->, line width=1pt] (sent-1-3) to [out=120,in=60, looseness=1]  (sent-1-1);
		\draw [chainLine, ->, line width=1pt] (sent-1-3) to [out=120,in=60, looseness=1]  (sent-1-2);
		\node [entboxc, above=of sent-1-4, text height=4mm, minimum width=4mm, xshift=0mm, yshift=-6mm, color=white] (c1)  [] {};
		\draw [chainLine, ->] (sent-1-3) to [out=60,in=120, looseness=1]  (c1.north);
		\draw [chainLine, ->] (sent-1-3) to [out=60,in=120, looseness=1]  (sent-1-5);
		\node [entboxc, above=of sent-1-6, text height=4mm, minimum width=4mm, xshift=0mm, yshift=-6mm,color=white] (c2)  [] {};
		\draw [chainLine, ->] (sent-1-5) to [out=60,in=120, looseness=1]  (c2.north);
		
		\draw [chainLine, ->, line width=1pt] (sent-1-8) to [out=120,in=60, looseness=1]  (sent-1-7);
		\draw [chainLine, ->, line width=1pt] (sent-1-8) to [out=120,in=60, looseness=1]  (sent-1-3);
		\draw [chainLine, ->] (sent-1-8) to [out=60,in=120, looseness=1.4]  (sent-1-9);
		\node [entboxc, above=of sent-1-14, text height=4mm, minimum width=4mm, xshift=0mm, yshift=-6mm,color=white] (c3)  [] {};
		\draw [chainLine, ->, line width=1pt] (sent-1-8) to [out=60,in=120, looseness=1]  (c3.north);
		\draw [chainLine, ->] (sent-1-9) to [out=60,in=120, looseness=1]  (sent-1-10);
		\draw [chainLine, ->] (sent-1-10) to [out=60,in=120, looseness=1.4]  (sent-1-12);
		\draw [chainLine, ->, line width=1pt] (sent-1-12) to [out=120,in=60, looseness=1.3]  (sent-1-11);
		\draw [chainLine, ->] (sent-1-12) to [out=60,in=120, looseness=1.2]  (sent-1-13);
		
		\node [below= of sent-1-10, yshift=-2mm, inner sep=1pt] (product) [] {\color{myred}\textbf{\textsc{Product}}};
		\draw [chainLine, ->, dashed, myred] (sent-1-13) to node[midway, below,align=center]{\textbf{?}}  (product);
		
		\begin{pgfonlayer}{background}
		\node [entbox, below=of sent-1-2, text height=8mm, minimum width=44mm, xshift=-3mm, yshift=6.3mm] (e1)  [] {\color{myred}\textbf{\textsc{Org}}};
		\node [entbox, below=of sent-1-13, text height=8mm, minimum width=10mm, yshift=6.3mm] (e1)  [] {\color{myred}\textbf{\textsc{Org}}};
		\end{pgfonlayer}

		
		\matrix (deppath) [matrix of nodes, nodes in empty cells, execute at empty cell=\node{\strut};, below=of sent, yshift=-8mm, xshift=-18.5mm]
        {	
			\textcolor{purple}{\textbf{Dependency Path:}} &[0mm]  \textbf{\textit{Corp.}} & [0mm] \textit{begin} & [0mm] \textit{trading} & [0mm] \textit{with} &[0mm]  \textit{symbol} & [0mm] \textbf{\textit{PCP}} \\
		};
		
		\matrix (hybpath) [matrix of nodes, nodes in empty cells, execute at empty cell=\node{\strut};, below=of sent, yshift=-22mm]
		{	
			\textcolor{purple}{\textbf{Hybrid Paths:}} &[-1mm] \textbf{\textit{Corp.}} & \textit{begin} &[-1mm]  \textit{\textbf{.}} & [0mm] \textbf{\textit{PCP}} & \textcolor{purple}{OR} &  \textbf{\textit{Corp.}} & [-1mm] \textit{begin} & [-1mm] \textit{trading} & [-1mm] \textit{with} &[-1mm] the &[-1mm] \textit{symbol} & [-1mm] \textbf{\textit{PCP}} \\
		};
		

		\draw [chainLine, -, line width=1pt] (deppath-1-2) to [out=60,in=120]  (deppath-1-3);
		\draw [chainLine, -, line width=1pt] (deppath-1-3) to [out=60,in=120]  (deppath-1-4);
		\draw [chainLine, -, line width=1pt] (deppath-1-4) to [out=60,in=120]  (deppath-1-5);
		\draw [chainLine, -, line width=1pt] (deppath-1-5) to [out=60,in=120]  (deppath-1-6);
		\draw [chainLine, -, line width=1pt] (deppath-1-6) to [out=60,in=120]  (deppath-1-7);
		
		\draw [chainLine, -, line width=1pt] (hybpath-1-2) to [out=60,in=120, line width=1pt]  (hybpath-1-3);
		\node [entboxc, above=of hybpath-1-4, text height=4mm, minimum width=4mm, xshift=-1.5mm, yshift=-6mm,color=white] (c4)  [] {};
		\draw [chainLine, -, line width=1pt] (hybpath-1-3) to [out=60,in=120]  (c4.north);
		\draw [chainLine, -, white] (c4.north east) to [out=60,in=120] node[myred, above, yshift=-1mm, xshift=1.5mm] {neighbor}  (hybpath-1-5);
		
		\draw [chainLine, -, line width=1pt] (hybpath-1-7) to [out=60,in=120]  (hybpath-1-8);
		\draw [chainLine, -, white] (hybpath-1-9) to [out=60,in=120] node[myred, above, yshift=-1.5mm, xshift=5mm] {context}  (hybpath-1-10);
		
		
		\end{tikzpicture} 
	}
	\caption{A sentence annotated with dependency trees and named entities. The paths to connect two entities are shown below the sentence.}
	\label{fig:intro}
\end{figure}
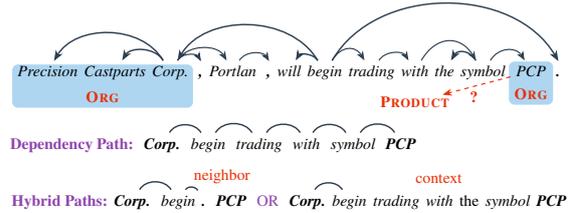

However, sequence models such as bidirectional LSTM~\cite{lstm} are not able to fully capture the long-range dependencies \cite{bengio2009learning}.
For instance, Figure \ref{fig:intro} (top) shows one type of structured information in NER.
The words ``{\it Precision Castparts Corp.}'' can be easily inferred as \textsc{Organization} by its context (i.e., \textit{Corp.}). 
{\color{black}However}, the second entity ``{\it PCP}'' could be misclassified as a \textsc{Product} entity if a model relies more on the context ``\textit{begin trading with}'' but ignores the hidden information that ``{\it PCP}'' is the symbol of ``{\it Precision Castparts Corp.}''. 

Previous research works~\cite{li2017leveraging,Jie2019DependencyGuidedLF,wang2019chinese} have been using the parse trees~\cite{chomsky,chomsky1969aspects,sandra2014morphological} to incorporate such structured information.
Figure \ref{fig:intro} (Dependency Path) 
shows that the first entity can be connected to the second entity following the dependency tree with 5 hops. 
Incorporating the dependency information can be done with graph neural networks (GNNs) such as graph convolutional networks (GCNs)~\cite{kipf2017semi}.
However, simply stacking the LSTM and GCN architectures for NER can only provide us with modest improvements; sometimes, it decreases performance~\cite{Jie2019DependencyGuidedLF}.
Based on the dependency path in Figure \ref{fig:intro}, it requires a 5-layer 
GCN to capture the connections between these two entities.
However, deep GCN architectures often face training difficulties, which cause a performance drop~\cite{Hamilton2017InductiveRL,kipf2017semi}.
Directly stacking GCN and LSTM has difficulties in modeling the interaction between dependency trees and contextual information.
\textcolor{black}{
To address the above limitations, we propose the Synergized-LSTM (Syn-LSTM), a new recurrent neural network architecture that considers an additional graph-encoded representation to update the memory and hidden states, as shown in Figure \ref{fig:Syn-LSTM}.
More specifically, the graph-encoded representation for each word can be obtained with GCNs. 
Our proposed Syn-LSTM allows the cell to receive the structured information from the graph-encoded representation. 
With the newly designed gating mechanism, our model is able to make independent assessments on the amounts of information to be retrieved from the word representation and the graph-encoded representation respectively. 
Such a mechanism allows for better integration of both contextual  and structured information.}

Our contributions can be summarized as:
\begin{itemize}
    \item We propose a \textcolor{black}{simple and robust} Syn-LSTM model to better incorporate the structured information conveyed by dependency trees.
    The output of the Syn-LSTM cell is jointly determined by both contextual  and structured information.
    We adopt the classic conditional random fields (CRF)~\cite{lafferty2001conditional} on top of the Syn-LSTM for NER.
    
    \item We conduct extensive experiments on several standard datasets across four languages.
    The proposed model significantly outperforms previous approaches on these datasets. 
    
    \item We show that the proposed model can capture long-distance interactions between entities. 
    Our further analysis  statistically demonstrates the proposed gating mechanism is able to aggregate the structured information selectively. 
\end{itemize}






\section{Synergized-LSTM}

\subsection{Incorporating Structured Information}
\begin{figure}
    \centering
\resizebox{1.0\linewidth}{!}{
\begin{tikzpicture}[
    font=\sf \scriptsize,
     >=Stealth,
    cell/.style={
        rectangle, 
        rounded corners=3mm, 
        draw,
        very thick,
        },
    operator/.style={
        circle,
        draw,
        inner sep=-0.5pt,
        minimum height =.2cm,
        line width=0.8pt
        },
    function/.style={
        ellipse,
        draw,
        inner sep=1pt
        },
    ct/.style={
        circle,
        draw,
        line width = 1pt,
        minimum width=7.5mm,
        inner sep=0.1pt,
        },
    gt/.style={
        rectangle,
        draw,
        minimum width=4mm,
        minimum height=3mm,
        inner sep=0.2pt
        },
    mylabel/.style={
        font=\scriptsize\sffamily
        },
    ArrowC1/.style={
        rounded corners=.1cm,
        thick,
        },
    ArrowC2/.style={
        rounded corners=.3cm,
        thick,
        },
    ]

    \node [cell, minimum height =5cm, minimum width=8cm, draw=none, fill=lowgreen!50] at (0,0){} ;

    \node [gt, minimum height=4mm] (fi) at (-2.1,-0.25) {\large $\sigma$};
    \node [gt, minimum height=4mm] (ig) at (0.4,-0.25) {\large$\sigma$};
    \node [gt, minimum height=4mm] (ix) at (-1.3,-0.25) {\large$\sigma$};
    \node [gt, minimum height=4mm] (cg) at (1.2,-0.25) {\large$tanh$};
    \node [gt, minimum height=4mm] (cx) at (-0.5,-0.25) {\large$tanh$};
    \node [gt, minimum height=4mm] (og) at (2.25,-0.25) {\large$\sigma$};

    \node [operator] (mux1) at (-2.1,2) {\Large $\times$};
    \node [operator] (addg) at (1.2,2) {\LARGE+};
    \node [operator] (muxg) at (1.2,0.8) {\Large$\times$};
    \node [operator] (addx) at (-.5,2) {\LARGE +};
    \node [operator] (muxx) at (-.5,0.8) {\Large$\times$};
    \node [operator] (mux3) at (2.75,0.45) {\large$\times$};
    \node [function] (func1) at (2.75,1.25) {\large$tanh$};

    \node[ct, label={[align=center] Previous \\ Cell}] (c) at (-5,2) {\fontsize{14}{14} $\mathbf{c}_{t\text{-}1}$};
    \node[ct, label={[align=center] Previous \\Hidden}] (h) at (-5,-1) {\fontsize{13}{13} $\mathbf{h}_{t\text{-}1}$};
    \node[ct, , color=skyblue, label={[align=center]left:Current \\Input}] (x) at (-3.5,-3.5) {\fontsize{18}{18} $\mathbf{x}_{t}$};
    \node[ct, , color=orange, label={[align=center]right:Graph-encoded \\Representation}] (g) at (-2.75,-3.5) {\fontsize{15}{15} $\mathbf{g}_{t}$};

    \node[mylabel, color=littledarkblue] (f) at (-2.3,0.2) {\fontsize{11}{11}\selectfont  $f_t$};
    \node[mylabel, color=littledarkblue] (f) at (-1.5,0.2) {\fontsize{11}{11}\selectfont $i_t$};
    \node[mylabel, color=littledarkblue] (f) at (-0.8,0.2) {\fontsize{11}{11}\selectfont $\tilde{{c}}_t$};
    \node[mylabel, color=littledarkblue] (f) at (0.13,0.2) {\fontsize{11}{11}\selectfont $m_t$};
    \node[mylabel, color=littledarkblue] (f) at (0.95,0.2) {\fontsize{11}{11}\selectfont $\tilde{{s}}_t$};
    \node[mylabel, color=littledarkblue] (f) at (2, 0.2) {\fontsize{11}{11}\selectfont $o_t$};

    \node [coordinate] (xx) at (-2.2, -1.5){};
    \node [coordinate] (gg) at (-2.2, -2){};

    \node[ct, label={[align=center]Current \\Cell}] (ct) at (5,2) {\fontsize{15}{15}  $\mathbf{c}_{t}$};
    \node[ct, label={[align=center]Current \\Hidden}] (ht) at (5,-2) {\fontsize{15}{15}  $\mathbf{h}_{t}$};
    \node[ct, label={[align=center]left:Current \\Hidden}] (hht) at (3.5,3.5) {\fontsize{15}{15}  $\mathbf{h}_{t}$};

    \draw [ArrowC1] (c) -- (mux1) -- (addx) -- (addg) -- (ct);
    

    \draw [ArrowC2, color=orange, line width=0.3mm] (g) |- (gg);
    \draw [ArrowC2, color=orange, line width=0.3mm] (gg) -| (fi);
    \draw [ArrowC2, color=orange, line width=0.3mm] (gg) -| (ig);
    \draw [ArrowC2, color=orange, line width=0.3mm] (gg) -| (cg);
    \draw [ArrowC2, color=orange, line width=0.3mm] (gg) -| (og);
    
    \draw [ArrowC2, color=skyblue, line width=0.3mm] (x) |- (xx);
    \draw [ArrowC2, color=skyblue, line width=0.3mm] (xx) -| (fi);
    \draw [ArrowC2, color=skyblue, line width=0.3mm] (xx) -| (ix);
    \draw [ArrowC2, color=skyblue, line width=0.3mm] (xx) -| (cx);
    \draw [ArrowC2, color=skyblue, line width=0.3mm] (xx) -| (og);

    

    

    \draw [ArrowC2] (h) -| (og);
    \draw [ArrowC2] (h) -| (fi);
    \draw [ArrowC2] (h) -| (ix);
    \draw [ArrowC2] (h) -| (ig);
    \draw [ArrowC2] (h) -| (cx);
    \draw [ArrowC2] (h) -| (cg);
    
    \draw [->, ArrowC1] (fi) -- (mux1);
    \draw [->, ArrowC1] (ix) |- (muxx);
    \draw [->, ArrowC1] (cx) -- (muxx);
    \draw [->, ArrowC1] (ig) |- (muxg);
    \draw [->, ArrowC1] (cg) -- (muxg);
    \draw [->, ArrowC1] (muxg) -- (addg);
    \draw [->, ArrowC1] (og) |- (mux3);
    \draw [->, ArrowC1] (muxx) -- (addx);
    \draw [->, ArrowC1] (addg -| func1)++(-0.5,0) -| (func1);
    \draw [->, ArrowC1] (func1) -- (mux3);

    \draw [-, ArrowC1] (mux3) |- (ht);
    \draw (ct -| hht) ++(0,-0.1) coordinate (i1);
    \draw [-, ArrowC1] (ht -| hht)++(-0.5,0) -| (i1);
    \draw [-, ArrowC1] (i1)++(0,0.2) -- (hht);
\end{tikzpicture}
}
\caption{Syn-LSTM cell. $t$ is the current time step.}
\label{fig:Syn-LSTM}
\end{figure}
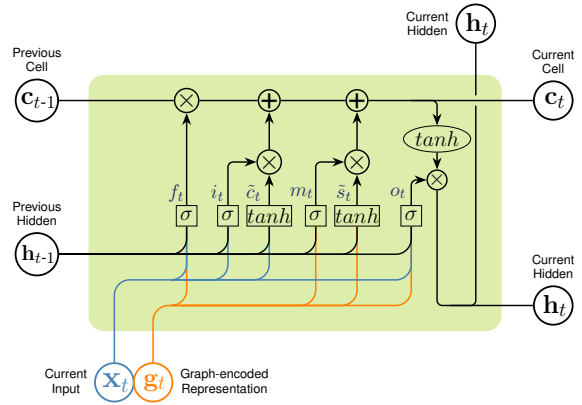

\textcolor{black}{
To incorporate the long-range dependencies, we consider an additional graph-encoded representation $\mathbf{g}_t$ (Figure \ref{fig:Syn-LSTM}) as the model input to integrate  contextual and structured information. 
The graph-encoded representation $\mathbf{g}_t$ can be derived from Graph Neural Networks (GNNs) such as GCN \cite{kipf2017semi}, which are capable of bringing in structured information through graph structure~\cite{Hamilton2017RepresentationLO}.
}

However, structured information sometimes is hard to encode, as we can see {\color{black}from the example in} Figure \ref{fig:intro}. 
One naive approach is to use a deep GNN to capture {\color{black}such information} along multiple dependency arcs between two words, which could mess up information and lead to training difficulties. 
A straightforward solution is to integrate both structured and contextual information via LSTM. 
As shown in Figure \ref{fig:intro} (Hybrid Paths), the structured information can be passed to neighbors or context, which allows a model to use less number of GNN layers and alleviate such issues for long-range dependencies.
The input to the LSTM can simply be the concatenation of word representation $\mathbf{x}_t$ and $\mathbf{g}_t$ at each position~\cite{Jie2019DependencyGuidedLF}\footnote{They concatenate the current word and head word representations.}.  
However, because such an approach requires both $\mathbf{x}_t$ and $\mathbf{g}_t$ to  decide the value of the input gate jointly, it could be a potential victim of two sources of uncertainties: 1) the uncertainty of the quality of graph-encoded representation $\mathbf{g}_t$, and 2) the uncertainty of the exact interaction mechanism between the two types of features. 
These may lead to sub-optimal performance, especially if the graph-encoded representation $\mathbf{g}_t$ is unsatisfactory.
Thus, we need to design a new approach to incorporate both types of information from $\mathbf{x}_t$ and $\mathbf{g}_t$ with a more explicit interaction mechanism, with which we hope to alleviate the above issues.

\subsection{Syn-LSTM Cell}
We propose the Synergized-LSTM (Syn-LSTM) to better integrate  the contextual and structured information to address the above limitations. 
\textcolor{black}{
The inputs of the Syn-LSTM cell include previous cell state $\mathbf{c}_{t-1}$, previous hidden state $\mathbf{h}_{t-1}$, current cell input $\mathbf{x}_t$, and an additional graph-encoded representation  $\mathbf{g}_t$. The outputs of the Syn-LSTM cell include current cell state $\mathbf{c}_t$ and current hidden state $\mathbf{h}_t$. Within the cell, there are four gates: input gate $\mathbf{i}_t$, forget gate $\mathbf{f}_t$, output gate $\mathbf{o}_t$, and an additional new gate  $\mathbf{m}_t$ to control the flow of information.
Note that the forget gate $\mathbf{f}_t$ and output gate $\mathbf{o}_t$ are not just looking at $\mathbf{h}_{t-1}$ and $\mathbf{x}_t$; they are also affected by the graph-encoded representation $\mathbf{g}_t$. 
}
The cell state $\mathbf{c}_t$ and hidden state $\mathbf{h}_t$ are computed as follows:
\begin{align}
    \label{forgetgate}
    \mathbf{f}_t &= \sigma(W^{(f)}\mathbf{x}_t + U^{(f)}\mathbf{h}_{t-1}  + Q^{(f)}\mathbf{g}_{t}  + \mathbf{b}^{(f)})\\ 
    \mathbf{o}_t &= \sigma(W^{(o)}\mathbf{x}_t + U^{(o)}\mathbf{h}_{t-1}  + Q^{(o)}\mathbf{g}_{t} + \mathbf{b}^{(o)})\\
    \mathbf{i}_t &= \sigma(W^{(i)}\mathbf{x}_t + U^{(i)}\mathbf{h}_{t-1} + \mathbf{b}^{(i)})\\
    \mathbf{m}_t &= \sigma(W^{(m)}\mathbf{g}_t + U^{(m)}\mathbf{h}_{t-1} + \mathbf{b}^{(m)})\\
    \tilde{\textbf{c}}_t &= \mathrm{tanh}(W^{(u)}\mathbf{x}_t + U^{(u)}\mathbf{h}_{t-1} + \mathbf{b}^{(u)})\\
    \tilde{\textbf{s}}_t &= \mathrm{tanh}(W^{(n)}\mathbf{g}_t + U^{(n)}\mathbf{h}_{t-1} + \mathbf{b}^{(n)})\\
    \label{cellstate}
    \mathbf{c}_t &= \mathbf{f}_t \odot \mathbf{c}_{t-1} + \mathbf{i}_t \odot \tilde{\textbf{c}}_t + \mathbf{m}_t \odot \tilde{\textbf{s}}_t  \\ 
    \mathbf{h}_t &= \mathbf{o}_t \odot \mathrm{tanh}(\mathbf{c}_t)
\end{align}
where $\sigma$ is the sigmoid function, $W^{(\cdot)}$, $U^{(\cdot)}$, $Q^{(\cdot)}$ and $\mathbf{b}^{(\cdot)}$ are learnable parameters.

\textcolor{black}{
The additional new gate $\mathbf{m}_t$ is used to  control the information from the graph-encoded representation directly. 
Such a design allows the original input gates $\mathbf{i}_t$ and our new gate $\mathbf{m}_t$ to make independent assessments on the amounts of information to be retrieved from the word representation $\mathbf{x}_t$ and the graph-encoded representation $\mathbf{g}_t$ respectively.
On the other hand, we also have a different candidate state $\tilde{\textbf{s}}_t$ to represent the cell state that corresponds to the graph-encoded representation separately.
}

With the proposed Syn-LSTM, 
the structured information captured by the dependency trees can be passed to each cell,
and the additional gate $\mathbf{m}_t$ is able to control how much structured information can be incorporated.
The additional gate enables the model to feed the contextual and structured information into the LSTM cell separately. 
Such a mechanism allows our model to aggregate the information from linear sequence and dependency trees selectively.

Similar to the previous work \cite{levy-etal-2018-long}, it is also possible to show that the cell state  $\mathbf{c}_t$  implicitly computes the element-wise weighted sum of the previous states by expanding Equation \ref{cellstate}:
\begin{align}
    \mathbf{c}_t &=
    \mathbf{f}_t \odot \mathbf{c}_{t-1} + \mathbf{i}_t \odot \tilde{\textbf{c}}_t + \mathbf{m}_t \odot \tilde{\textbf{s}}_t  \\\nonumber
    &=  \sum_{j=0}^t ( \mathbf{i}_j \odot \prod_{k=j+1}^t \mathbf{f}_k ) \odot \tilde{\textbf{c}}_j \\ 
    &+ \sum_{j=0}^t ( \mathbf{m}_j \odot \prod_{k=j+1}^t \mathbf{f}_k ) \odot \tilde{\textbf{s}}_j\\
    & = \sum_{j=0}^t \textbf{a}_j^t \odot \tilde{\textbf{c}}_j + \sum_{j=0}^t \textbf{q}_j^t \odot \tilde{\textbf{s}}_j
\end{align}

Note that the two terms, $\textbf{a}_j^t$ and $\textbf{q}_j^t $ , are the product of gates. The value of the two terms are in the range from 0 to 1.
Since the $\tilde{\textbf{c}}_t$ and $\tilde{\textbf{s}}_t$ represent contextual and structured features, the corresponding weights control the flow of information.

\begin{figure}[t!]
		\centering
		\adjustbox{max width=0.9\linewidth}{
			\begin{tikzpicture}[node distance=8.0mm and 10mm, >=Stealth, 
			crf/.style={
				draw=fontgray,
				circle,
				double,
				minimum height=10mm, 
				minimum width=10mm,
				line width=0.7pt, 
				inner sep=1pt, 
			},
			lstm/.style={
				rectangle,
				rounded corners=1mm,
				fill=lowgreen!50,
				minimum height=7mm, 
				minimum width=13mm, 
				label={center:\scriptsize \textcolor{fontgray}{Syn-LSTM}}
			},
			chainLine/.style={
				line width=0.8pt,->, 
				color=mygray
			},
			embg/.style={
				circle,
				draw,
				color=orange,
				line width = .8pt,
				minimum width=7mm,
				text width=5mm,
				inner sep = 0.1
			},
			gcnnode/.style={
				circle,
				draw=lightorange,
				line width = .8pt,
				minimum width=7mm,
				text width=5mm,
				inner sep = 0.1
			},
			embx/.style={
				circle,
				draw,
				color = skyblue,
				line width = .8pt,
				minimum width=7mm,
				text width=5mm,
				inner sep=0.1
			},
			entbox/.style={draw=none, rounded corners, fill=myyellow} 
			]
			
			
			\node[embx] (x1) {\small $\mathbf{x}_{t\text{-}1}$};
			\node[embx, right= of x1, xshift=6mm, align=center] (x2) {\small $\mathbf{x}_{t}$};
			\node[embx, right= of x2, xshift=6mm] (x3) {\small $\mathbf{x}_{t\text{+}1}$};
			\node[embx, right= of x3, xshift=6mm] (x4) {\small $\mathbf{x}_{t\text{+}2}$};
			
			\node[embg, right= of x1, xshift=-8mm] (g1) {\small $\mathbf{g}_{t\text{-}1}^{L}$};
			\node[embg, right= of x2, xshift=-8mm, align=center] (g2) {\small $\mathbf{g}_{t}^L$};
			\node[embg, right= of x3, xshift=-8mm] (g3) {\small $\mathbf{g}_{t\text{+}1}^L$};
			\node[embg, right= of x4, xshift=-8mm] (g4) {\small $\mathbf{g}_{t\text{+}2}^L$};
			
			\node[lstm, above= of x1, xshift=4.5mm](m1) {};
			\node[lstm, right= of m1](m2) {};
			\node[lstm, right= of m2](m3) {};
			\node[lstm, right= of m3](m4) {};

			\node[crf, above= of m1, yshift=0mm] (e1) { \textcolor{black}{$y_{t-1}$}};
			\node[crf, above= of m2, yshift=0mm](e2) { \textcolor{black}{$y_t$}};
			\node[crf, above= of m3, yshift=0mm](e3) { \textcolor{black}{$y_{t+1}$}};
			\node[crf,above= of m4, yshift=0mm](e4) { \textcolor{black}{$y_{t+2}$}};
			
			\draw [chainLine, ->]  (g1) to (m1);
			\draw [chainLine, ->] (g2) -> (m2);
			\draw [chainLine, ->] (g3) -> (m3);
			\draw [chainLine, ->] (g4) -> (m4);
			\draw [chainLine, ->] (x1) -> (m1);
			\draw [chainLine, ->] (x2) -> (m2);
			\draw [chainLine, ->] (x3) -> (m3);
			\draw [chainLine, ->] (x4) -> (m4);
			
			\draw [chainLine, <->] (m1) to (m2);
			\draw [chainLine, <->] (m2) to (m3);
			\draw [chainLine, <->] (m3) to (m4);
			
			\draw [chainLine, ->] (m1) -> (e1);
			\draw [chainLine, ->] (m2) -> (e2);
			\draw [chainLine, ->] (m3) -> (e3);
			\draw [chainLine, ->] (m4) -> (e4);
			
			\draw [chainLine,  -, middlefactor] (e1) to (e2);
			\draw [chainLine,  -, middlefactor] (e2)to (e3);
			\draw [chainLine,  -, middlefactor] (e3)to (e4);

			\node[embg, below= of g1, xshift=0mm, yshift=-8mm] (g01) {\small $\mathbf{g}_{t\text{-}1}^{0}$};
			\node[embg, below= of g2, xshift=0mm, yshift=-8mm] (g02) {\small $\mathbf{g}_{t}^0$};
			\node[embg, below= of g3, xshift=0mm, yshift=-8mm] (g03) {\small $\mathbf{g}_{t\text{+}1}^0$};
			\node[embg, below= of g4, xshift=0mm, yshift=-8mm] (g04) {\small $\mathbf{g}_{t\text{+}2}^0$};
			
			\draw [chainLine,  ->] (g01) to (g1);
			\draw [chainLine,  ->] (g02) to (g2);
			\draw [chainLine,  ->] (g03) to (g3);
			\draw [chainLine,  ->] (g04) to (g4);
			
			\node [entbox, below right=of g2, text height=5mm, minimum width=80mm, xshift=-41mm, yshift=2mm, text centered, label={[anchor=north, inner sep=3pt, fontgray]center: \small Graph Convolutional Network}] (gcn) {};
			

			
			\end{tikzpicture} 
		}
		\caption{Syn-LSTM-CRF architecture. 
		}
		\label{fig:archi}
	\end{figure}
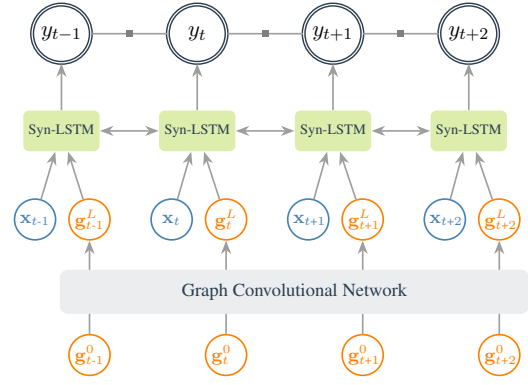

\section{Syn-LSTM-CRF}
The goal of named entity recognition is to predict the label sequence $\mathbf{y} = \{y_1,y_2, ..., y_n \}$  given the input sequence $\mathbf{w} = \{w_1,w_2, ..., w_n \}$, where  $w_t$ represents the $t$-th word and $n$ is the number of words. 
Our model is mainly constructed with three layers: input representation layer, bi-directional Syn-LSTM layer, and CRF layer. 
The architecture of our Syn-LSTM-CRF is shown in Figure \ref{fig:archi}.

\paragraph{Input Representation Layer}
Similar to the work by \citet{lample2016neural}, our input representation also includes the character embeddings, which are the hidden states of character-based BiLSTM. \citet{Jie2019DependencyGuidedLF} highlight that the dependency relation helps to enhance the input representation.
Furthermore, previous methods \cite{wang2018nner, wang2018oner} use embeddings of part-of-speech (POS) tags as additional input representation. The input representation $\mathbf{x}_t$ of our model is the concatenation of the word embedding $\mathbf{v}_t$, the character representation $\mathbf{e}_t$, the dependency relation embedding $\mathbf{r}_t$, and the POS embedding $\mathbf{p}_t$:
\begin{equation}
\mathbf{x}_t= [ \mathbf{v}_t; \;\mathbf{e}_t; \; \mathbf{r}_t; \; \mathbf{p}_t]
\end{equation}
where both $\mathbf{r}_t$ and $\mathbf{p}_t$ embeddings are randomly initialized and are fine-tuned during training. For experiments with the contextualized representations (e.g., BERT~\cite{devlin2019bert}), we further concatenate the contextual word representation to $\mathbf{x}_t$.

For our task, we employ the graph convolutional network \cite{kipf2017semi, zhang2018graph} to get the graph-encoded representation $\mathbf{g}_t$. Given a graph, an adjacency matrix $A$ of size  $n \times n$ is able to represent the graph structure, where  $n$ is the number of nodes; $A_{i,j} = 1$ indicates that node $i$ and node $j$ are connected. We transform  dependency tree into its corresponding adjacency matrix\footnote{We treat the dependency edge as undirected and add a self-loop for each node: $A_{i,j}=A_{j,i}$ and $A_{i,i} =1$.} $A$,  and $A_{i,j} = 1$ denotes that node $i$ and node $j$ have dependency relation. Note that the purpose of graph-encoded representation $\mathbf{g}_t$ is to incorporate the dependency information from  neighbor nodes. The input and output representations of the $l$-th layer GCN  at $t$-th position are denoted as $\mathbf{g}_t^{l-1}$  and  $\mathbf{g}_t^l$ respectively. Similar to the work by \citet{zhang2018graph}, we  use $d_t = \sum_{j=1}^{n}A_{t,j}$, which is the total number of  neighbors of node $t$, to normalize the representation before going through the nonlinear function. The GCN operation is defined as:
\begin{align}
    \mathbf{g}_t^l = \mathrm{ReLU}(\sum_{j=1}^{n} A_{t,j} W^{l}\mathbf{g}_t^{l-1}/d_t + \mathbf{b}^{l})
\end{align}
\textcolor{black}{where $W^l$ is a linear transformation and $\mathbf{b}^l$ is a bias.}
The initial $\mathbf{g}_t^0$ is the concatenation of word embedding $\mathbf{v}_t$, character embedding $\mathbf{e}_t$, and dependency relation embedding $\mathbf{r}_t$: $	\mathbf{g}_t^0= [ \mathbf{v}_t; \;\mathbf{e}_t; \;\mathbf{r}_t]$.


\paragraph{Bi-directional Syn-LSTM Layer}
With the word representation $\mathbf{x}_t$ and the graph-encoded representation $\mathbf{g}_t$, a bi-directional Syn-LSTM is applied to generate contextual representation. The forward and backward Syn-LSTM enable the model to integrate the contextual and structured  information from both directions. We concatenate the hidden state $\overrightarrow{\mathbf{h}_t}$ from forward Syn-LSTM and hidden state $\overleftarrow{\mathbf{h}_t}$ from backward Syn-LSTM to form the contextual representation of $t$-th token: $\mathbf{h}_t = [\overrightarrow{\mathbf{h}_t} ; \; \overleftarrow{\mathbf{h}_t}]$.

\paragraph{CRF Layer}
The CRF~\cite{lafferty2001conditional} is widely used in NER tasks as it is capable of capturing the structured correlations between adjacent output labels. 
Given the sentence $\mathbf{w}$ and dependency tree $\mathbf{\tau}$, the probability of the label sequence $\mathbf{y}$ is defined as:
\begin{equation}
P(\mathbf{y} \vert \mathbf{w}, \mathbf{\tau})
=
\frac
{\exp ( score(\mathbf{w}, \mathbf{\tau}, \mathbf{y})  )  }
{\sum_{\mathbf{y}^\prime }\exp ( score(\mathbf{w}, \mathbf{\tau}, \mathbf{y}^\prime) )  }
\end{equation}

The score function is defined as:
\begin{equation}
score(\mathbf{w}, \mathbf{\tau} , \mathbf{y}) = \sum_{t=0}^n T_{y_t, y_{t+1}} + \sum_{t=1}^n E_{y_t}
\end{equation}
where $T_{y_t, y_{t+1}}$ denotes the transition score from label $y_t$ to $y_{t+1}$, $E_{y_t}$ denotes the score of label $y_t$ at the $t$-th position and the scores are computed using the hidden state $\mathbf{h}_t$.
We learn the model parameters by minimizing the negative log-likelihood and employ the Viterbi algorithm to obtain the best label sequence during evaluation.

\begin{table}[t!]
	\centering
	\resizebox{1\linewidth}{!}{
	\begin{tabular}{llrrrrrr}
		\toprule
		\multirow{2}{*}{\textbf{Dataset}}& & \multirow{2}{*}{\textbf{\# Sent.}} & \multicolumn{5}{c}{\textbf{\# Entity in Sentence  Length}} \\\cmidrule(lr){4-8} 
		 &&& $\mathbf{\leq}$ 14 & 15 - 29 & 30 - 44 & 45 - 59  & $\mathbf{\geq}$ 60  \\
		 \midrule
		 \multirow{3}{*}{\textbf{Catalan}} 
         & \textbf{Train} & 8,709&  944 & 4,821 & 5,309 & 2,815 &1,389 \\
		 & \textbf{Dev} & 1,445&135 & 836 & 815 & 477 & 168\\
         & \textbf{Test} & 1,698 & 243 & 919 & 946 & 518 & 284 \\
		 \midrule
		 \multirow{3}{*}{\textbf{Spanish}}
         & \textbf{Train} & 9,022& 855 & 4,031 & 6,656 & 4,279 & 1,446\\
		 & \textbf{Dev} & 1,419&  125 & 612 & 911 & 707 & 260 \\
		 & \textbf{Test} & 1,705& 175 & 703 & 1,143 & 783 & 242 \\\midrule
		  \multirow{3}{*}{\textbf{English}} 
         &\textbf{Train} &  59,924  & 13,309 & 33,853 &22728 & 8,099 & 3,839 \\
		 &\textbf{Dev} &  8,528 & 1,778 & 4,830 & 2,882 & 1,051 & 525  \\ 
		 &\textbf{Test} & 8,262 &  1,785 & 4,673 & 3,171 & 1,082 & 546 \\
		 \midrule
		 \multirow{3}{*}{\textbf{Chinese}} 
         & \textbf{Train} &36,487 & 8,424 & 21,033 & 17,260 & 8,392 & 7,434\\ 
		 & \textbf{Dev} & 6,083& 1,493 & 3,250 & 2,284& 1,099 & 978 \\
		 & \textbf{Test} & 4,472&  968 & 2,517 & 2,149 & 1,024 & 836\\
		 \bottomrule
	\end{tabular}
	}
	\caption{Statistics of datasets.}
	\label{tab:datastat}
\end{table}

\section{Experiments}
\label{sec:results}
\paragraph{Datasets}
The proposed model is evaluated on four benchmark datasets:  SemEval 2010 Task 1
~\cite{recasens2010semeval} Catalan and Spanish datasets, and OntoNotes 5.0 \cite{weischedel2013ontonotes} English and Chinese datasets. 
\textcolor{black}{We choose these four datasets as they have explicit dependency annotations which allow us to evaluate the effectiveness of our approach when dependency trees of different qualities are used.}
For SemEval 2010 Task 1 datasets, there are 4 entity types: \textsc{Per}, \textsc{Loc} and \textsc{Org} and \textsc{Misc}.
For OntoNotes 5.0 datasets, there are 18 entity types in total. 
Following the work by \citet{Jie2019DependencyGuidedLF}, we transform the parse trees into the Stanford dependency trees \cite{de2008stanford} by using Stanford CoreNLP \cite{manning-etal-2014-stanford}. 
Detailed statistics of each dataset can be found in Table \ref{tab:datastat}.
Intuitively, longer sentences would require the model to capture more long-distance interactions in the sentences.
We present the number of entities in terms of different sentence lengths to show that these datasets have a modest amount of entities in long sentences.

\paragraph{Experimental Setup}
For Catalan, Spanish, and Chinese,  we use the FastText \cite{grave2018learning} 300 dimensional embeddings to initialize the word embeddings. 
For OntoNotes 5.0 English, we adopt the publicly available GloVE \cite{pennington2014glove} 100 dimensional embeddings to initialize the word embeddings.  
For experiments with the contextualized representation, we adopt the pre-trained language model BERT \cite{devlin2019bert} for the four datasets.  
Specifically, we use bert-as-service \cite{xiao2018bertservice} to generate the contextualized word representation without fine-tuning. 
Following \citet{Luo2019HierarchicalCR}, we use the cased version of BERT large model for the experiments on the OntoNotes 5.0 English data. 
We use the cased version of BERT base model for the experiments on the other three datasets.
For the character embedding, we randomly initialize the character embeddings and set the dimension as 30, and set the hidden size of character-level BiLSTM as 50. 
The hidden size of GCN and Syn-LSTM is set as 200, the number of GCN layer is 2. 
We adopt stochastic gradient descent (SGD) to optimize our model with batch size 100, L2 regularization $10^{-8}$, {\color{black}initial} learning rate $lr$ 0.2 and the learning rate is decayed\footnote{We set the decay as 0.1 and the learning rate for each epoch equals to $lr/(1 + decay*(epoch-1))$.} with respect to the number of epoch. 
We select the best model based on the performance on the dev set\footnote{The experimental results on the dev set and other experimental details can be found in the Appendix.} and apply it to the test set. 
We use the bootstrapping t-test to compare the results.

\paragraph{Baselines} 
We compare our model with several baselines with or without dependency tree information. 
The first one is BERT-CRF, where we apply a CRF layer on top of BERT~\cite{devlin2019bert}. 
Secondly, we compare with the BERT implementation by HuggingFace~\cite{Wolf2019HuggingFacesTS}. 
For models with dependency trees, we take the {\color{black}models} BiLSTM-GCN-CRF and dependency-guided LSTM-CRF (DGLSTM-CRF) proposed by \citet{Jie2019DependencyGuidedLF}, and our implemented GCN-BiLSTM-CRF. 
The BiLSTM-GCN-CRF {\color{black}model} simply stacks the GCN on top of the BiLSTM to incorporate the dependency trees. 
The GCN-BiLSTM-CRF {\color{black}model} takes the concatenation of the graph-encoded representation from GCN and word embedding as input into BiLSTM.
{\color{black}The DGLSTM-CRF takes the concatenation of the head word representation and word embedding as input into BiLSTM.
Note that the original implementation of DGLSTM-CRF uses ELMo~\cite{peters2018deep}, {\color{black}but} we also implement it with BERT. 
Besides, we compare our model with previous works that have results on these datasets. }


\begin{table}[!t]
    \centering
    
	\setlength{\tabcolsep}{4pt} 
	\renewcommand{\arraystretch}{1} 
    \resizebox{1\linewidth}{!}{
    \begin{tabular}{lcccccc}
    \toprule
    \multirow{2}{*}{\textbf{Models}} & \multicolumn{3}{c}{\textbf{Catalan}}& \multicolumn{3}{c}{\textbf{Spanish}} \\
    \cmidrule(lr){2-4}\cmidrule(lr){5-7}
     & $P.$ & $R.$ & $F_1$ & $P.$ & $R.$ & $F_1$  \\ \midrule
            BiLSTM-CRF$^{\dag}$ & 76.83 & 63.47 & 69.51 &78.33 & 69.89 & 73.87 \\
        	BiLSTM-GCN-CRF$^{\dag}$ & 81.25 & 75.22 & 78.12 &84.10 & 79.88 & 81.93\\
        	GCN-BiLSTM-CRF$^*$ & 80.95&	74.19&	77.43& 84.36&	79.48&	81.85\\
			DGLSTM-CRF (\citeyear{Jie2019DependencyGuidedLF})& 83.35 & 80.00 & 81.64& 84.05 & 82.90 &83.47\\
            Syn-LSTM-CRF (Ours)& \textbf{83.90} & \textbf{81.65} & \textbf{82.76}    & \textbf{86.22} & \textbf{84.24} & \textbf{85.09}\\ \midrule
            \multicolumn{3}{l}{\textbf{+ Contextualized Word Representation}} &\\
            BERT-CRF$^*$& 76.34 & 76.05 & 76.19 & 79.30 & 77.22 & 78.24 \\
            \citet{Wolf2019HuggingFacesTS}$^*$ & 82.82 &	85.7 &	84.23 & 81.36 &	85.58 &	83.42 \\ 
            BiLSTM-CRF $_{\scriptscriptstyle \text{+ ELMO}}$$^{\dag}$ &77.85 & 76.22 & 77.03&81.72 & 79.09 & 80.38 \\
            BiLSTM-CRF $_{\scriptscriptstyle \text{+ BERT}}$$^*$ &81.21 & 79.90 & 80.55 &83.28 & 80.11 & 81.66 \\
            BiLSTM-GCN-CRF$_{\scriptscriptstyle \text{+ ELMO}}$$^{\dag}$ &83.68 & 83.16 & 83.42& 85.31 & 85.19 & 85.25\\
            GCN-BiLSTM-CRF$_{\scriptscriptstyle \text{+ BERT}}$$^*$ & 87.60 &	86.39 &	86.99 & 88.07 &	87.46 &	87.76\\ 
            DGLSTM-CRF (\citeyear{Jie2019DependencyGuidedLF})$_{\scriptscriptstyle \text{+ ELMO}}$ & 84.71 & 83.75 & 84.22 & 87.79 & 87.33 & 87.56  \\
            DGLSTM-CRF$_{\scriptscriptstyle \text{+ BERT}}$$^*$ &  85.92 & 84.50 & 85.20 &   85.67 & 85.00 & 85.33 \\
            Syn-LSTM-CRF$_{\scriptscriptstyle \text{+ BERT}}$ (Ours)  & \textbf{89.07} & \textbf{89.04} & \textbf{89.05 }&   \textbf{89.66} & \textbf{90.54} & \textbf{90.10} \\
    \bottomrule
    \end{tabular}
    }
    \caption{Experimental results [\%] on SemEval 2010 Task 1 Catalan and Spanish test set. The models with * symbol are our implementations. The models with $^{\dag}$ symbol are retrieved from \citet{Jie2019DependencyGuidedLF}. }
    \label{tab:semeval_results}
\end{table}

\subsection{Main Results}
\paragraph{SemEval 2010 Task 1}
\label{sec:semeval}
Table \ref{tab:semeval_results} shows comparisons of our model with baseline models on 
the SemEval 2010 Task 1 Catalan and Spanish datasets.
Our Syn-LSTM-CRF model outperforms all existing models with $F_1$ 82.76 and 85.09 ($p < 10^{-5}$) compared to DGLSTM-CRF on Catalan and Spanish datasets when FastText word embeddings are used. 
Our model outperforms the BiLSTM-CRF model by 13.25 and 11.22 $F_1$ points, and outperforms BiLSTM-GCN-CRF \cite{Jie2019DependencyGuidedLF} model by 4.64 and 3.16 on Catalan and Spanish. 
The large performance gap between BiLSTM-GCN-CRF and our model indicates that Syn-LSTM-CRF shows better compatibility with GCN, and this confirms that simply stacking GCN on top of the BiLSTM does not perform well. 
Our method outperforms GCN-BiLSTM-CRF model by 5.33 and 3.24 $F_1$ points on Catalan and Spanish. This shows that our proposed model demonstrates a better integration of  contextual information and structured information.
Furthermore, our proposed method brings 1.12 and 1.62 $F_1$ points improvement on  Catalan and Spanish datasets compare to the DGLSTM-CRF \cite{Jie2019DependencyGuidedLF}. 
The DGLSTM-CRF employs 2-layer dependency guided BiLSTM to capture grandchild dependencies, which leads to longer training time and more model parameters. However, our Syn-LSTM-CRF is able to get better performance with fewer model parameters and shorter training time because of the fewer LSTM layers. Such results demonstrate that our proposed Syn-LSTM-CRF manages to  capture structured information effectively.

Furthermore, with the contextualized word representation, the Syn-LSTM-CRF$_{\scriptscriptstyle \text{+ BERT}}$ achieves much higher performance improvement than any other method. 
Our model outperforms the strong baseline model DGLSTM-CRF$_{\scriptscriptstyle \text{+ ELMO}}$ by 4.83 and 2.54 in terms of $F_1$  ($p < 10^{-5}$) on Catalan and Spanish, respectively. 
\begin{table}[!t]
    \centering
    \resizebox{1.0\linewidth}{!}{
    \begin{tabular}{lccc}
    \toprule
     \textbf{Models} &  $P.$ & $R.$ & $F_1$ \\ \midrule
     		\citet{chiu2016named} & 86.04 & 86.53 & 86.28\\
			\citet{li2017leveraging}   & 88.00 & 86.50 & 87.21 \\
			\citet{strubell2017fast} & - & - &  86.84 \\
			\citet{ghaddar2018robust} & - & - & 87.95 \\
            BiLSTM-CRF$^{\dag}$ & 87.21 & 86.93 & 87.07 \\
            BiLSTM-GCN-CRF$^{\dag}$ &88.30 & 88.06 & 88.18 \\
            GCN-BiLSTM-CRF$^*$ & 88.56&	88.76&	88.66\\
            DGLSTM-CRF (\citeyear{Jie2019DependencyGuidedLF}) & 88.53 & 88.50& 88.52 \\ 
            \citet{Luo2019HierarchicalCR} &  - & - & 87.98 \\
            
            Syn-LSTM-CRF (Ours) &  \textbf{88.96} & \textbf{89.13} & \textbf{89.04} \\\midrule
            \multicolumn{3}{l}{\textbf{+ Contextualized Word Representation}} &\\
            \citet{akbik2018coling}&- & - & 89.30 \\
            BERT-CRF$^*$ & 88.42 & 88.33 & 88.37\\
            \citet{Wolf2019HuggingFacesTS}$^*$ & 88.39 & 90.29&	89.33\\
            BiLSTM-CRF$_{\scriptscriptstyle \text{+ ELMO}}$$^{\dag}$   & 89.14 & 88.59 & 88.87 \\
            BiLSTM-CRF$_{\scriptscriptstyle \text{+ BERT}}$$^*$   &  89.32	& 90.02 &	89.67 \\
            BiLSTM-GCN-CRF$_{\scriptscriptstyle \text{+ ELMO}}$$^{\dag}$  & 89.40 & 89.71 & 89.55 \\
            GCN-BiLSTM-CRF$_{\scriptscriptstyle \text{+ BERT}}$$^*$  & 89.34	& 91.26 &	90.29\\
            DGLSTM-CRF (\citeyear{Jie2019DependencyGuidedLF})$_{\scriptscriptstyle \text{+ ELMO}}$ & 89.59 & 90.17 & 89.88 \\
            DGLSTM-CRF$_{\scriptscriptstyle \text{+ BERT}}$$^*$  &89.63&	89.87&	89.75 \\
            \citet{Luo2019HierarchicalCR}$_{\scriptscriptstyle \text{+ BERT}}$   & - & - & 90.30 \\
            Syn-LSTM-CRF$_{\scriptscriptstyle \text{+ BERT}}$ (Ours) & 	\textbf{90.14}&	\textbf{91.58}& \textbf{90.85}\\
    \bottomrule
    \end{tabular}
    }
    \caption{Experimental results [\%] on OntoNotes 5.0 English test set. The models with * symbol are our implementations.  The models with $^{\dag}$ symbol are retrieved from \citet{Jie2019DependencyGuidedLF}. There are also other methods \cite{Li2020AUM, Li2020DiceLF} that use external information, \cite{yu-etal-2020-named} use document-level information to encode the sentence,  which are not direct comparisons to ours.  }
    \label{tab:english_results}
\end{table}
\paragraph{OntoNotes 5.0 English}
\label{sec:english}

To understand the generalizability of our model, we evaluate the proposed Syn-LSTM-CRF model on large scale OntoNotes 5.0 datasets. Table \ref{tab:english_results} shows comparisons of our model with baseline models on English. Our Syn-LSTM-CRF model outperforms all existing methods with 89.04 in terms of $F_1$ score ($p < 0.01$) compared to DGLSTM-CRF, when GloVE word embeddings are used. 
Our model outperforms the BiLSTM-CRF model by 1.97 in $F_1$, BiLSTM-GCN-CRF \cite{Jie2019DependencyGuidedLF} model by 0.86. Note that our implemented GCN-BiLSTM-CRF outperforms the previous  DGLSTM-CRF \cite{Jie2019DependencyGuidedLF} by 0.14 in $F_1$. 
Our Syn-LSTM-CRF further brings the improvement to 0.52.
Moreover, with the contextualized word representation BERT, our method achieves an $F_1$ score of 90.85 ($p < 10^{-5}$) compared to DGLSTM-CRF$_{\scriptscriptstyle \text{+ ELMO}}$. Our method outperforms the previous model \cite{Luo2019HierarchicalCR}, which relies on document-level information, by 0.55 in $F_1$. Furthermore, the performance improvement on recall is more prominent as compared to precision. 
This shows that the proposed Syn-LSTM-CRF is able to extract more entities.

\begin{table}[!t]
    \centering
    \resizebox{1.0\linewidth}{!}{
    \begin{tabular}{lccc}
    \toprule
    \textbf{Models} & $P.$ & $R.$ & $F_1$ \\ \midrule
 			\citet{pradhan2013towards} &78.20 & 66.45 & 71.85\\ 
			Lattice LSTM (\citeyear{zhang-yang-2018-chinese})&76.34& 77.01& 76.67 \\
            BiLSTM-CRF$^{\dag}$&\textbf{78.45} & 74.59 & 76.47\\
            BiLSTM-GCN-CRF$^{\dag}$& 76.35 & 75.89 & 76.12 \\
            GCN-BiLSTM-CRF$^*$ & 78.30 &	75.07&	76.65 \\
            DGLSTM-CRF (\citeyear{Jie2019DependencyGuidedLF}) &77.40 & 77.41 & 77.40\\
            Syn-LSTM-CRF (Ours)  & 77.95 & \textbf{79.07} & \textbf{78.51} \\ \midrule
            \multicolumn{3}{l}{\textbf{+ Contextualized Word Representation}} &\\
            BERT-CRF$^*$& \textbf{79.83 }& 79.68 & 79.75\\
            \citet{Wolf2019HuggingFacesTS}$^*$ & 77.35 &	81.74&	79.49\\
            BiLSTM-CRF$_{\scriptscriptstyle \text{+ ELMO}}$$^{\dag}$ &79.20 & 79.21 & 79.20 \\
            BiLSTM-CRF$_{\scriptscriptstyle \text{+ BERT}}$$^*$ &78.45 & 81.24 & 79.82 \\
            BiLSTM-GCN-CRF$_{\scriptscriptstyle \text{+ ELMO}}$$^{\dag}$ &  78.71 & 79.29 & 79.00\\
            GCN-BiLSTM-CRF$_{\scriptscriptstyle \text{+ BERT}}$$^*$ & 79.03 &	80.98&	80.00\\
            DGLSTM-CRF (\citeyear{Jie2019DependencyGuidedLF})$_{\scriptscriptstyle \text{+ ELMO}}$ & 78.86 & 81.00 & 79.92 \\
            DGLSTM-CRF$_{\scriptscriptstyle \text{+ BERT}}$$^*$  & 77.79 & 81.65 & 79.67  \\
            Syn-LSTM-CRF$_{\scriptscriptstyle \text{+ BERT}}$ (Ours) &  78.66 & \textbf{81.80} & \textbf{80.20}\\
    \bottomrule
    \end{tabular}
    }
    \vspace{-1mm}
    \caption{Experimental results [\%] on OntoNotes 5.0 Chinese test set. The models with * symbol are our implementations. The models with $^{\dag}$ symbol are retrieved from \citet{Jie2019DependencyGuidedLF}. There are also other methods \cite{Li2020AUM, Li2020DiceLF} that use external information, which are not direct comparisons to ours.  }
    \label{tab:chinese_results}
    \vspace{-2mm}
\end{table}

\paragraph{OntoNotes 5.0 Chinese }
We present the experimental results on the OntoNotes 5.0 Chinese test set in  Table \ref{tab:chinese_results}. 
Our model still consistently outperforms the baseline models, specifically by 2.04 in $F_1$ compared to BiLSTM-CRF, by 2.39 compared to  BiLSTM-GCN-CRF, by 1.86 compared to GCN-BILSTM-CRF and by 1.11 ($p < 10^{-5}$) compared to DGLSTM-CRF when FastText  is used. Note that the baseline BiLSTM-GCN-CRF model is 0.35 points worse than  BiLSTM-CRF. 
Such results further confirm the effectiveness of our proposed Syn-LSTM-CRF for incorporating structured information. We find a similar behavior when the contextualized word representation BERT is used. With the contextualized word representation, we achieve a higher $F_1$ score of 80.20.

\begin{table*}[t!]
	\centering
	\resizebox{1\linewidth}{!}{
		\begin{tabular}{lccc|ccc|ccc|ccc}
			\toprule
			\multirow{2}{*}{\textbf{Models}} & \multicolumn{3}{c}{\textbf{Catalan}}& \multicolumn{3}{c}{\textbf{Spanish}}  & \multicolumn{3}{c}{\textbf{English}}& \multicolumn{3}{c}{\textbf{Chinese}}\\
            \cmidrule(lr){2-4}\cmidrule(lr){5-7}\cmidrule(lr){8-10}\cmidrule(lr){11-13}
            & $P.$ & $R.$ & $F_1$ & $P.$ & $R.$ & $F_1$   & $P.$ & $R.$ & $F_1$ & $P.$ & $R.$ & $F_1$  \\ \midrule
            DGLSTM-CRF$_{\scriptscriptstyle \text{+ ELMO}}$ (Given) & 84.71 &	83.75 &84.22& 87.79& 87.33 & 87.56 & 89.59 & 90.17 & 89.88 & 78.86 &	81.00 &79.92 \\
		    DGLSTM-CRF$_{\scriptscriptstyle \text{+ ELMO}}$ (Predicted)  & -& -& 82.37&-&- & 83.92 & -& - & 89.64 & -& -& 79.59 \\
		    \textbf{Differences} & -& -& \textbf{-1.85}&-&- & \textbf{-3.64} & -& - & \textbf{-0.24} & -& -& \textbf{-0.33} \\
		    DGLSTM-CRF$_{\scriptscriptstyle \text{+ ELMO}}$ (Random) & 78.99 & 79.31 & 79.15 & 82.11 & 80.89 & 81.49 & 88.80 & 88.91 & 88.85 & 77.68 & 80.60 & 79.11\\
		    \textbf{Differences } & \textbf{-5.72}& \textbf{-4.44}& \textbf{-5.07}&\textbf{-5.68}&\textbf{-6.44} & \textbf{-6.07}& \textbf{-0.79} & \textbf{-1.26} & \textbf{-1.03}& \textbf{-1.18}& \textbf{-0.40} & \textbf{-0.81} \\ \midrule
		    Syn-LSTM-CRF$_{\scriptscriptstyle \text{+ BERT}}$ (Given) & 89.07 & 89.04 & 89.05 & 89.66 & 90.54 &90.10 & 90.14 & 91.58 & 90.85 & 78.66 & 81.80& 80.20  \\ 
		    Syn-LSTM-CRF$_{\scriptscriptstyle \text{+ BERT}}$ (Predicted) &87.33 & 87.42 & 87.38 & 86.50 & 87.49 & 86.99 & 89.91 & 91.27 & 90.58 & 78.86 & 81.57 &80.19  \\
		    \textbf{Differences} & \textbf{-1.74}& \textbf{-1.62}& \textbf{-1.67}&\textbf{-3.16}&\textbf{-3.05} & \textbf{-3.11} & \textbf{-0.23}& \textbf{-0.31} & \textbf{-0.27} & \textbf{+0.20} & \textbf{-0.23}& \textbf{-0.01}\\
		    Syn-LSTM-CRF$_{\scriptscriptstyle \text{+ BERT}}$ (Random) & 84.57 &	85.53 &	85.05 & 84.61 &	86.61 &	85.59 & 89.24 & 90.46 & 89.84 & 77.25	& 81.91 &	79.51\\
		    \textbf{Differences} & \textbf{-4.50}& \textbf{-3.51}& \textbf{-4.00}&\textbf{-5.05} & \textbf{-3.93} & \textbf{-4.51}&\textbf{ -0.90} & \textbf{-1.12} & \textbf{-1.01}& \textbf{-1.41}& \textbf{-0.11} &\textbf{-0.69}\\
			\bottomrule
		\end{tabular}
	}
	\vspace*{-2mm}
	\caption{Performance comparison between adopting the given, predicted and random dependencies on SemEval 2010 Task 1 Catalan and Spanish, and OntoNotes 5.0 English and Chinese datasets. Note that DGLSTM-CRF$_{\scriptscriptstyle \text{+ ELMO}}$ have better performance compared to  DGLSTM-CRF$_{\scriptscriptstyle \text{+ BERT}}$ based on Table \ref{tab:semeval_results}, \ref{tab:english_results}, \ref{tab:chinese_results}. }
	\vspace*{-2mm}
	\label{tab:dep_result}
\end{table*}


\section{Analysis}
\paragraph{Robustness  Analysis}
\textcolor{black}{
To study the robustness of our model and check whether our model can regulate the flow of information from the graph-encoded representation, we analyze the influence of the quality of dependency trees. We train and evaluate an additional dependency parser \cite{dozat2017deep}. Specifically, we train the dependency parser\footnote{The performance of the dependency parser can be found in the Appendix.} on the given training datasets and select the best model based on the dev sets. Then we apply the best model to the test sets to obtain dependency trees. We also train and evaluate our model with random dependency trees. 
Table \ref{tab:dep_result} presents the comparisons between Syn-LSTM-CRF$_{\scriptscriptstyle \text{+ BERT}}$ and DGLSTM-CRF$_{\scriptscriptstyle \text{+ ELMO}}$ with given, predicted and random dependency trees.
We observe that both models encounter a performance drop when we use the predicted parse trees and random trees. 
Our performance differences with the given parse trees are relatively smaller than the corresponding differences in DGLSTM-CRF$_{\scriptscriptstyle \text{+ ELMO}}$.
Such an observation demonstrates the robustness of our proposed model against structured information from the trees of different quality.
It is worthwhile to note that, with the predicted dependencies, our proposed Syn-LSTM-CRF$_{\scriptscriptstyle \text{+ BERT}}$ is still able to outperform the strong baseline DGLSTM-CRF$_{\scriptscriptstyle \text{+ ELMO}}$ even with the given parse trees on Catalan, English, and Chinese datasets.}

\begin{figure}[t!]
\centering
\begin{tikzpicture}[scale=0.67]
\pgfplotsset{width=6.5cm, height=5cm, compat=1.3}
\begin{axis}[
    xtick={1,2,3,4,5,6,7,8,9,10},
    ymin=0, ymax=2.4e7,
    xticklabels = {0-0.4, 0.4-0.5, 0.5-0.6, 0.6-0.7,0.7-0.8,0.8-0.9,0.9-1},
    xticklabel style = {font=\fontsize{9}{1}\selectfont, rotate=40, xshift=2mm},
    yticklabel style = {font=\fontsize{9}{1}\selectfont, },
    legend style={font=\fontsize{6.5}{1}\selectfont},
	y label style={at={(axis description cs:-0.12,.5)},anchor=south, xshift=2cm},
	legend style={at={(0.5,0.75)},anchor=south,legend columns=1}, 
	label style={font=\fontsize{9}{1}\selectfont},
	ybar=2pt,
	bar width=5pt,
    ]
\addplot[orange,pattern=north west lines, pattern color=orange,  area legend] coordinates {
(1, 39896) (2, 9101407) (3, 15813498) (4,931705) (5,19617) (6, 990)(7,2)
 };
 \addplot[skyblue, pattern=grid, pattern color=skyblue, area legend]  coordinates {
(1, 351284) (2, 5138439) (3, 10033093) (4,3976201) (5,1826827) (6, 2496932)(7, 1719760)
};
\legend{ With Random Parse Tree\\ With Given Parse Tree\\}
\end{axis}
\end{tikzpicture}
\begin{tikzpicture}[scale=0.67]
\pgfplotsset{width=6.5cm, height=5cm, compat=1.3}
\begin{axis}[
    xtick={1,2,3,4,5,6,7,8,9,10},
    ymin=0, ymax=2.4e7,
    xticklabels = {0-0.4, 0.4-0.5, 0.5-0.6, 0.6-0.7,0.7-0.8,0.8-0.9,0.9-1},
    xticklabel style = {font=\fontsize{9}{1}\selectfont, rotate=40, xshift=2mm},
    yticklabel style = {font=\fontsize{9}{1}\selectfont, },
    legend style={font=\fontsize{6.5}{1}\selectfont},
	y label style={at={(axis description cs:-0.12,.5)},anchor=south, xshift=2cm},
	legend style={at={(0.5,0.75)},anchor=south,legend columns=1}, 
	label style={font=\fontsize{9}{1}\selectfont},
	ybar=2pt,
	bar width=5pt,
    ]
\addplot[orange,pattern=north west lines, pattern color=orange,  area legend] coordinates {
(1, 27995) (2, 4576287) (3, 14490610) (4,6227154) (5,825625) (6, 22311)(7,172)
 };
 \addplot[skyblue, pattern=grid, pattern color=skyblue, area legend]  coordinates {
(1, 276270) (2, 4394214) (3, 8628870) (4,4584658) (5,3787709) (6, 3307130)(7, 879563)
};
\legend{With Random Parse Tree\\With Given Parse Tree\\}
\end{axis}
\end{tikzpicture}
\caption{Left: Catalan, Right: Spanish. 
$x$-axis: the value of gate $\mathbf{m}_t$. $y$-axis: the number of words.
}
\label{fig:gating}
\end{figure}
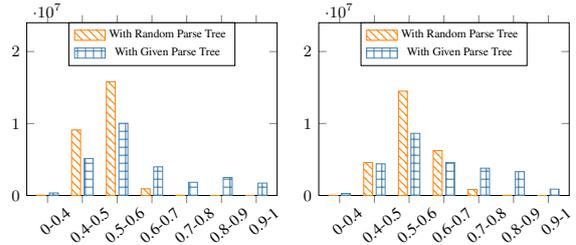

\textcolor{black}{
To further study the robustness, we conduct an analysis to investigate if the gate $\mathbf{m_t}$ (Figure \ref{fig:Syn-LSTM}) has the  ability to regulate the flow of information from the graph-encoded representation.
Intuitively, the gate $\mathbf{m}_t$ {\color{black}should tend} to have a small value when the quality of the parse tree is not good (e.g., with random trees). 
We statistically plot the number of words with respect to different gate value ranges ($\mathbf{m}_t$).
Figure \ref{fig:gating} shows the comparison between the models of using random trees and given trees on Catalan and Spanish\footnote{We found a similar behavior for OntoNotes 5.0 English and Chinese datasets, and the detailed result can be found in the Appendix.}. 
We observe that the gate $\mathbf{m}_t$ is more likely to open (the value is higher) when we use the given parse trees compared with random parse trees. 
Such behavior demonstrates that our proposed model can selectively aggregate the information from the graph-encoded representation.}

\paragraph{Effect of Sentence Length}

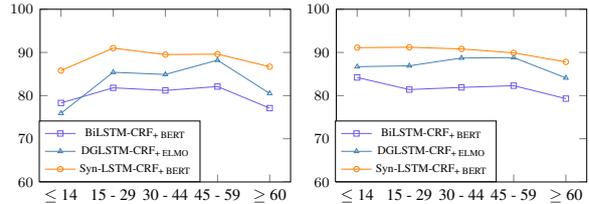
\begin{figure}[t!]
\centering
\begin{tikzpicture}[scale=0.67]
\pgfplotsset{width=6.5cm, height=5cm, compat=1.3}
\begin{axis}[
    xtick={1,2,3,4,5},
    ymin=60, ymax=100,
    xticklabels = {$\leq$ 14, 15 - 29 ,30 - 44 , 45 - 59 , $\geq$ 60},
    xticklabel style = {font=\fontsize{9}{1}\selectfont, rotate=0, xshift=0mm},
    yticklabel style = {font=\fontsize{8}{1}\selectfont, },
    legend style={font=\fontsize{6.5}{1}\selectfont},
	y label style={at={(axis description cs:-0.12,.5)},anchor=south, xshift=2cm},
	legend style={at={(0.64,0.355)}},
	label style={font=\fontsize{9}{1}\selectfont},
    ]
    \addplot [mark=square, mark size=1.5pt, color=fruitpurple] plot coordinates {
    (1, 78.3) (2, 81.8) (3, 81.2) (4, 82.1) (5, 77.1) };
    \addplot [mark=triangle, mark size=1.5pt, color=skyblue] plot coordinates {
    (1, 75.9) (2, 85.4) (3, 84.9) (4, 88.2) (5, 80.5) };
    \addplot [mark=o, mark size=1.5pt, color=orange] plot coordinates {
    (1, 85.8) (2, 91.0) (3, 89.5) (4, 89.6) (5, 86.7) };
\legend{ BiLSTM-CRF$_{\scaleto{\text{+ BERT}}{3.5pt}}$\\ DGLSTM-CRF$_{\scaleto{\text{+ ELMO}}{3.5pt}}$\\ Syn-LSTM-CRF$_{\scaleto{\text{+ BERT}}{3.5pt}}$\\}
\end{axis}
\end{tikzpicture}
\begin{tikzpicture}[scale=0.67]
\pgfplotsset{width=6.5cm, height=5cm, compat=1.3}
\begin{axis}[
    xtick={1,2,3,4,5},
    ymin=60, ymax=100,
    xticklabels = {$\leq$ 14, 15 - 29 ,30 - 44 , 45 - 59 , $\geq$ 60},
    xticklabel style = {font=\fontsize{9}{1}\selectfont, rotate=0, xshift=0mm},
    yticklabel style = {font=\fontsize{8}{1}\selectfont, },
    legend style={font=\fontsize{6.5}{1}\selectfont},
	y label style={at={(axis description cs:-0.12,.5)},anchor=south, xshift=2cm},
	legend style={at={(0.64,0.355)}}, 
	label style={font=\fontsize{9}{1}\selectfont},
    ]
    \addplot [mark=square, mark size=1.5pt, color=fruitpurple] plot coordinates {
    (1, 84.2) (2, 81.4) (3, 81.9) (4, 82.3) (5, 79.3) };
    \addplot [mark=triangle, mark size=1.5pt, color=skyblue] plot coordinates {
    (1, 86.7) (2,86.9) (3, 88.7) (4, 88.8) (5, 84.1) };
    \addplot [mark=o, mark size=1.5pt, color=orange] plot coordinates {
    (1, 91.1) (2, 91.2) (3, 90.8) (4, 89.9) (5, 87.8) };
\legend{ BiLSTM-CRF$_{\scaleto{\text{+ BERT}}{3.5pt}}$\\ DGLSTM-CRF$_{\scaleto{\text{+ ELMO}}{3.5pt}}$\\ Syn-LSTM-CRF$_{\scaleto{\text{+ BERT}}{3.5pt}}$\\}
\end{axis}
\end{tikzpicture}
\caption{Left: Catalan, Right: Spanish. 
$x$-axis: sentence length. $y$-axis:$F_1$ score (\%). Note that DGLSTM-CRF$_{\scriptscriptstyle \text{+ ELMO}}$ have better performance compared to  DGLSTM-CRF$_{\scriptscriptstyle \text{+ BERT}}$ based on Table \ref{tab:semeval_results}, \ref{tab:english_results}, \ref{tab:chinese_results}.
}
\label{fig:sentencelength}
\end{figure}

We compare the performance of our Syn-LSTM-CRF$_{\scriptscriptstyle \text{+ BERT}}$ with BiLSTM-CRF$_{\scriptscriptstyle \text{+ BERT}}$ and DGLSTM-CRF$_{\scriptscriptstyle \text{+ ELMO}}$ models with respect to sentence length, and the results are shown in Figure \ref{fig:sentencelength}. 
We observe that the Syn-LSTM-CRF$_{\scriptscriptstyle \text{+ BERT}}$ model consistently outperforms the two baseline models on the four languages\footnote{See the Appendix for the results on  OntoNotes 5.0 English and Chinese datasets.}. 
In particular, although the performance tends to drop as the sentence length increases, our proposed model shows relatively better performance when the sentence length is $\geq 60$.  This confirms that the proposed Syn-LSTM-CRF$_{\scriptscriptstyle \text{+ BERT}}$ is able to effectively incorporate structured information. Note that our 2-layer GCN is computed based on the dependency trees, which include both short-range dependencies and long-range dependencies. With the graph-encoded representation and the proposed Syn-LSTM-CRF$_{\scriptscriptstyle \text{+ BERT}}$, the individual word representation is enhanced by both contextual and structured information. Therefore, for the sentences with length of $\leq 14$, we can still observe obvious improvements. 
The significant performance improvements on the four datasets show the capability of our Syn-LSTM-CRF to capture the structured information despite the sentence length.

\paragraph{Effect of Entity Length}
We conduct another evaluation on BiLSTM-CRF$_{\scriptscriptstyle \text{+ BERT}}$, DGLSTM-CRF$_{\scriptscriptstyle \text{+ ELMO}}$, and Syn-LSTM-CRF$_{\scriptscriptstyle \text{+ BERT}}$ models with respect to entity length $\in \{1,2,3,4,5, \geq6 \}$ on the four languages. Table \ref{tab:reslength} shows the performance comparison of two models with respect to entity length. With the structured information, both DGLSTM-CRF$_{\scriptscriptstyle \text{+ ELMO}}$ and Syn-LSTM-CRF$_{\scriptscriptstyle \text{+ BERT}}$ achieve better performance compared to BiLSTM-CRF$_{\scriptscriptstyle \text{+ BERT}}$. When the length of entity is $\leq$ 3, Syn-LSTM-CRF$_{\scriptscriptstyle \text{+ BERT}}$ achieves better results compared to DGLSTM-CRF$_{\scriptscriptstyle \text{+ ELMO}}$. This confirms that our proposed method can effectively incorporate the structured information. Our model consistently outperforms BiLSTM-CRF$_{\scriptscriptstyle \text{+ BERT}}$, and the performance tends to have more improvements when entities are getting longer except on the Chinese dataset. 
We note there are some special characteristics of the Chinese language.  
As mentioned by \citet{Jie2019DependencyGuidedLF}, 
the percentage of entities that are able to perfectly form a sub-tree is only $92.9\%$ for {\color{black}OntoNotes} Chinese, as compared to $98.5\%$, $100\%$, $100\%$ for OntoNotes English, SemEval Catalan and Spanish. 
Furthermore, the ratio
of long entities is much {\color{black}higher} for Catalan and Spanish compared to English and Chinese. The experimental results on Catalan and Spanish datasets show significant improvements for long entities. 
\textcolor{black}{Such results show that the structured information conveyed by the dependency trees can be more crucial when entity length becomes longer.}

\begin{table}[t!]
	\centering
	\resizebox{1\linewidth}{!}{
	\begin{tabular}{clcccccc}
		\toprule
		\multirow{2}{*}{\textbf{Dataset}}& \multirow{2}{*}{\textbf{Model}} & \multicolumn{6}{c}{\textbf{Entity Length}} \\
		 & & \textbf{1} & \textbf{2} & \textbf{3} & \textbf{4} & \textbf{5}  & $\mathbf{\geq}$\textbf{6}  \\
 		 \midrule
		 \multirow{2}{*}{\textbf{Catalan}} & BiLSTM-CRF$_{\scriptscriptstyle \text{+ BERT}}$ &82.4 & 84.4 & 77.8 & 53.3 & 31.8 & 36.2\\
		 & DGLSTM-CRF$_{\scriptscriptstyle \text{+ ELMO}}$ & 85.4 & 85.1 & 84.1 & \textbf{78.9} & 60.9 & 59.3\\
         & Syn-LSTM-CRF$_{\scriptscriptstyle \text{+ BERT}}$ &  \textbf{90.5} & \textbf{91.1} & \textbf{87.2} & 77.8 & \textbf{63.8} & \textbf{60.6}\\
		 \midrule
		 \multirow{2}{*}{\textbf{Spanish}} & BiLSTM-CRF$_{\scriptscriptstyle \text{+ BERT}}$ &85.1  & 84.2 & 81.5 & 33.7 & 43.1 & 27.2\\
		 & DGLSTM-CRF$_{\scriptscriptstyle \text{+ ELMO}}$ & 89.3 & 87.4 & 90.8 & \textbf{74.1} & 67.7 & \textbf{64.4}\\
		 & Syn-LSTM-CRF$_{\scriptscriptstyle \text{+ BERT}}$ & \textbf{92.7} & \textbf{90.9} & \textbf{91.1} & 73.0 & \textbf{75.4} & 58.5 \\
		 \midrule
		 \multirow{2}{*}{\textbf{English}} & BiLSTM-CRF$_{\scriptscriptstyle \text{+ BERT}}$ & 92.9 & 88.3 & 83.1 & 85.5 & 80.5 & 77.9 \\
		 & DGLSTM-CRF$_{\scriptscriptstyle \text{+ ELMO}}$ & 91.8 &  90.1 &  85.4& 87.0 &  \textbf{80.8} &  78.7 \\ 
		 & Syn-LSTM-CRF$_{\scriptscriptstyle \text{+ BERT}}$ &  \textbf{92.9} & \textbf{90.8} & \textbf{87.7} & \textbf{87.4} & 80.6 & \textbf{79.8}\\
		 \midrule
		 \multirow{2}{*}{\textbf{Chinese}} & BiLSTM-CRF$_{\scriptscriptstyle \text{+ BERT}}$ &82.5 & 74.6 & 71.4 & 65.0 &\textbf{69.8} & \textbf{52.5 }\\
		 & DGLSTM-CRF$_{\scriptscriptstyle \text{+ ELMO}}$ & 82.2& 75.5 & 71.8 & 64.1 & 58.5 & 41.1\\
		 & Syn-LSTM-CRF$_{\scriptscriptstyle \text{+ BERT}}$ & \textbf{82.5} & \textbf{75.6} & \textbf{73.1} & \textbf{66.4} & 66.1 & 42.5 \\

		 \bottomrule
	\end{tabular}
	}
    \caption{$F_1$ score [\%] based on entity length on Catalan, Spanish, English and Chinese datasets.  Note that DGLSTM-CRF$_{\scriptscriptstyle \text{+ ELMO}}$ have better performance compared to  DGLSTM-CRF$_{\scriptscriptstyle \text{+ BERT}}$ based on the results in the main paper.}
	\label{tab:reslength}
\end{table}

\begin{figure}[t!]
\centering
\begin{tikzpicture}[scale=0.89]
\pgfplotsset{width=7.5cm, height=5.5cm, compat=1.16}
\begin{axis}[
    xtick={1,2,3,4, 5},
    ymin=70, ymax=97,
    xticklabels = { Catalan, Spanish, English, Chinese, AVG},
    xticklabel style = {font=\fontsize{10}{1}\selectfont},
    yticklabel style = {font=\fontsize{10}{1}\selectfont},
    legend style={font=\fontsize{8}{1}\selectfont},
	ylabel={Dev $F_1$ $[\%]$},
	y label style={at={(axis description cs:-0.12,.5)},anchor=south},
	enlargelimits=0.14,
	legend style={at={(0.5,0.8)},anchor=south,legend columns=3}, 
	x label style={font=\fontsize{10}{1}\selectfont},
	y label style={font=\fontsize{10}{1}\selectfont},
	ybar=2pt,
	bar width=5pt,
    ]
\addplot[fruitpurple,pattern=north west lines, pattern color=fruitpurple,  area legend] coordinates {
(1, 89.28)(2, 89.28) (3, 89.08)(4, 78.49) (5, 86.04)
 };
 \addplot[orange, pattern=crosshatch, pattern color=orange, area legend] coordinates {
(1, 89.95) (2, 89.95) (3, 89.18) (4, 78.50)  (5, 86.6)
 };
 \addplot[skyblue, pattern=grid, pattern color=skyblue, area legend]  coordinates {
(1, 89.77)(2, 89.77) (3, 89.04)  (4,78.58) (5, 86.15)
};
\legend{ 1 layer\\2 layer\\3 layer\\}
\end{axis}
\end{tikzpicture}
\caption{Performance of different number of layers of GCN on dev set. }
\label{fig:gcnlayers}
\end{figure}
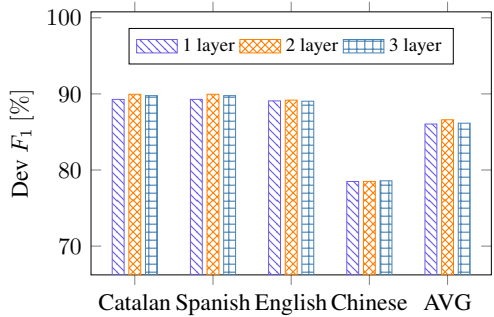

\paragraph{Number of GCN Layers}
To fully explore the impact of the number of GCN layers, we conduct another experiment on Syn-LSTM-CRF$_{\scriptscriptstyle \text{+ BERT}}$ model with the number of GCN layers $\in \{ 1,2,3\}$, and Figure \ref{fig:gcnlayers} shows the performance on the dev set of the four languages. The last bar, indicated as AVG, is obtained by averaging the dev results on the four datasets. We observe that the overall performance is better when the number of GCN layers equals 2. Note that similar behavior can also be found in the work by \citet{kipf2017semi} for document classification and node classification. Therefore, we evaluate our proposed Syn-LSTM-CRF model with 2-layer GCN.

\paragraph{Ablation Study}
To understand the contribution of each component,
we conduct an ablation study on the OntoNotes 5.0 English dataset, and Table \ref{tab:ablation} presents the detailed results  of our model with contextualized representation. We find that the performance drops by 0.24 $F_1$ score when we only use 1-layer GCN. Without GCN at all, the score drops by 1.13 $F_1$. The original dependency contributes 0.27 $F_1$ score. Removing the dependency relation embedding also decreases the performance by 0.27 $F_1$. When we remove the POS tags embedding, the result drops by 0.39 $F_1$.

\begin{table}[t!]
	\centering
	\resizebox{0.93\linewidth}{!}{
		\begin{tabular}{lccc}
			\toprule
			\textbf{Model}& $P.$ & $R.$ &   $F_\mathbf{1}$\\
			\midrule
			Syn-LSTM-CRF$_{\scriptscriptstyle \text{+ BERT}}$ & \textbf{90.14} & \textbf{91.58} & \textbf{90.85} \\
			~~~~-- $1$ layer GCN & 89.93 &	91.30 & 90.61\\
			~~~~-- $2$ layer GCN & 89.50 & 89.93 & 89.72 \\
			~~~~-- original dependency & 89.91 & 91.27 & 90.58\\
			~~~~-- dependency embedding& 89.85 & 91.31 & 90.58\\
			~~~~-- POS embedding &89.84 & 90.95 & 90.46\\

			
			\bottomrule
		\end{tabular}
	}
	\vspace{-1mm}
	\caption{Ablation study of the Syn-LSTM-CRF$_{\scriptscriptstyle \text{+ BERT}}$ model on OntoNotes 5.0 English. -- means removing.}
	\vspace{-1mm}
	\label{tab:ablation}
\end{table}

\section{Related Work}
\paragraph{LSTM}
LSTM has demonstrated its great effectiveness in many NLP tasks and becomes a standard module for many state-of-the-art models \cite{Wen2015SemanticallyCL, Ma2016EndtoendSL, dozat2017deep}. However, the sequential nature of the LSTM makes it challenging to capture long-range dependencies. \citet{Zhang2018SentenceStateLF} propose the S-LSTM model to include a sentence state to allow both local and global information exchange simultaneously. Mogrifier LSTM \cite{Moglstm2020} mutually gates the current input and the previous output to enhance the interaction between the input and the context. 
These two works do not consider structured information for the LSTM design.
Since natural language is usually structured, \citet{onlstm2018OrderedNI} propose ON-LSTM to add a hierarchical bias to allow the neurons to be updated by following {\color{black}certain} order. 
While the ON-LSTM is learning the latent constituency parse trees, we focus on incorporating the {\color{black}explicit} structured information conveyed by the dependency parse trees. 
\paragraph{NER}
Early work~\cite{sasano2008japanese} uses syntactic dependency features to improve the SVM performance on Japanese NER task. \citet{liu2010recognizing} propose to construct skip-edges to link similar words or words having typed dependencies to capture long-range dependencies. The later works~\cite{collobert2011natural,lample2016neural, chiu-nichols-2016-named} focus on using neural networks to extract features and achieved the state-of-the-art performance. \citet{jie2017efficient} find that some relations between the dependency edges and the entities {\color{black}can be used to} reduce the search space of their model, which significantly reduces the time complexity. \citet{yu-etal-2020-named} employ pre-trained language model to encode document-level information to explore  all spans with  the graph-based dependency graph based ideas. The  pre-trained language models (e.g., BERT~\cite{devlin2019bert}, ELMO~\cite{peters2018deep}) further improve neural-based approaches with a good contextualized representation. 
{\color{black}However, previous works did not focus on investigating how to effectively integrate structured and contextual information well.}

\section{Conclusion}
\textcolor{black}{
In this paper, we propose a simple and robust Syn-LSTM model to better integrate the structured information leveraged from the long-range dependencies.
Specifically, we introduce an additional graph-encoded representation to each recurrent unit. 
Such a graph-encoded representation can be obtained via GNNs. 
Through the newly designed gating mechanism, the hidden states are enhanced by contextual information captured by the linear sequence and structured information captured by the dependency trees. 
We present the Syn-LSTM-CRF for NER and adopt the GCN on dependency trees to obtain the graph-encoded representations.
Our extensive experiments and analysis on the datasets with four languages demonstrate that the proposed Syn-LSTM is able to effectively incorporate both contextual and structured information. The robustness analysis demonstrates that our model is capable of selectively aggregating the information from the graph-encoded representation.
}

\section*{Acknowledgements}

We would like to thank the anonymous reviewers for their helpful comments.
This research is partially supported by Ministry of Education, Singapore, under its Academic Research Fund (AcRF) Tier 2 Programme (MOE AcRF Tier 2 Award No: MOE2017-T2-1-156). 
Any opinions, findings and conclusions or recommendations expressed in this material are those of the authors and do not reflect the views of the Ministry of Education, Singapore.




\bibliography{custom}
\bibliographystyle{acl_natbib}

\appendix

\section{Experimental details}
We test our model on RTX 2080 Ti GPU and Nvidia Tesla V100 GPU, with CUDA version 10.1, PyTorch version 1.40. 
The average run time for Syn-LSTM is 52 sec/epoch,  55 sec/epoch,  290 sec/epoch,  350 sec/epoch for Catalan, Spanish, Chinese and English datasets respectively.
The total number of parameters is 11M.
Table \ref{tab:dev_results} shows the performance of our model on the dev sets of  OntoNotes 5.0 English and Chinese, SemEval 2010 Task 1 Catalan and Spanish.

For hyper-parameter, we use the FastText \cite{grave2018learning} 300 dimensional embeddings to initialize the word embeddings for Catalan, Spanish, and Chinese. 
For OntoNotes 5.0 English, we adopt the publicly available GloVE \cite{pennington2014glove} 100 dimensional embeddings to initialize the word embeddings.  
For experiments with the contextualized representation, we adopt the pre-trained language model BERT \cite{devlin2019bert} for the four datasets.  
Specifically, we use bert-as-service \cite{xiao2018bertservice} to generate the contextualized word representation without fine-tuning. 
Following \citet{Luo2019HierarchicalCR},
we select the $18^{th}$ layer of the cased version of BERT large model for the experiments on the OntoNotes 5.0 English data. 
We use the  the $9^{th}$ layer of cased version of BERT base model for the experiments on the rest three datasets.
For the character embedding, we randomly initialize the character embeddings and set the dimension as 30, and set the hidden size of character-level BiLSTM as 50. 
The hidden size of GCN and Syn-LSTM is set as 200. 
Note that we only use one layer of bi-directional Syn-LSTM for our experiments.
Dropout is set to 0.5 for input embeddings and hidden states. 
We adopt stochastic gradient descent (SGD) to optimize our model with batch size 100, L2 regularization $10^{-8}$, learning rate 0.2 and the learning rate is decayed with respect to the number of epoch \footnote{We set the decay as 0.1 and the learning rate for each epoch equals to $learning\_rate/(1 + decay*(epoch-1))$.} .

\begin{table}[t!]
	\centering
	\resizebox{1\linewidth}{!}{
		\begin{tabular}{lcccccccccccc}
			\toprule
			& \textbf{English} & \textbf{Chinese} & \textbf{Catalan} & \textbf{Spanish} \\
			\midrule
			Dependency LAS$\dagger$ & 94.89 & 89.28 & 93.25 & 93.35\\
			\bottomrule
		\end{tabular}
	}
	\caption{Performance of the trained dependency parser.}
	\label{tab:dep_result}
\end{table}

\begin{table}[t!]
	\centering
	\resizebox{1.0\linewidth}{!}{
	\begin{tabular}{llrrrrrr}
		\toprule
		\multirow{2}{*}{\textbf{Dataset}}& & \multicolumn{6}{c}{\textbf{Entity Length}} \\
		 & & \textbf{1} & \textbf{2} & \textbf{3} & \textbf{4} & \textbf{5}  & $\mathbf{\geq}$\textbf{6}  \\
		 \midrule
		 \multirow{3}{*}{\textbf{English}} 
         & Train & 46,525 & 17,391 & 9,714 & 4,892 & 1,938 & 1,368\\
         & Dev & 6,325 & 2,395 & 1,256 & 643 & 275 & 172 \\
         & Test & 6,129 & 2,598 & 1,359 & 706 & 278 & 187 \\
		 \midrule
		 \multirow{3}{*}{\textbf{Chinese}} 
         & Train & 47,285 & 9,668 & 3,626 & 1,139 & 467 & 358\\
         & Dev &  6,969 & 1,397 & 473 & 169 & 55 & 41\\
         & Test &  5,479 & 1299 & 473 & 146 & 55 & 42\\
		 \midrule
		 \multirow{3}{*}{\textbf{Catalan}} 
         & Train & 8,819 & 3,897 & 1,742 & 264 & 119 & 437\\
         & Dev &  1,370 & 676 & 269 & 40 & 18 & 58\\
         & Test & 1,601 & 811 & 338 & 57 & 27 & 76\\
		 \midrule
		 \multirow{3}{*}{\textbf{Spanish}} 
         & Train & 10,307 & 3,609 & 2,302 & 301 & 175 & 603 \\
         & Dev &  1,523 & 559 & 348 & 54 & 31 & 100\\
         & Test & 1,755 & 702 & 369 & 59 & 34 & 127\\
		 \bottomrule
	\end{tabular}
	}
    \caption{Number of entities with respect to entity length for OntoNotes 5.0 English and Chinese, SemEval 2010 Catalan and Spanish datasets.}
	\label{tab:statslength}
\end{table}

\begin{table*}[!t]
    \centering
    \resizebox{0.8\textwidth}{!}{
    \begin{tabular}{lccc|ccc|ccc|ccc}
    \toprule
    \multirow{2}{*}{\textbf{Models}} & \multicolumn{3}{c}{English} & \multicolumn{3}{c}{Chinese} & \multicolumn{3}{c}{Catalan}& \multicolumn{3}{c}{Spanish} \\ \cmidrule(lr){2-4}\cmidrule(lr){5-7}\cmidrule(lr){8-10}\cmidrule(lr){11-13}
     &  $P.$ & $R.$ & $F_1$ & $P.$ & $R.$ & $F_1$ &  $P.$ & $R.$ & $F_1$ & $P.$ & $.R$ & $F_1$\\ \midrule

            Syn-LSTM-CRF & 86.73 & 87.71 & 87.22 & 77.25 & 75.74 & 76.49 & 84.48 & 82.60 & 83.53 & 83.76 & 82.22 & 82.98\ \\ [1.5mm]
            Syn-LSTM-CRF$_{\scriptscriptstyle \text{+ BERT}}$  & 88.10&	90.27&	89.17&  78.05 & 78.84 & 78.45 & 89.87 & 89.76 & 89.81 & 88.50 & 88.60 & 88.55 \\
    \bottomrule
    \end{tabular}
    }
    \caption{Experimental results [\%] on dev set.}
    \label{tab:dev_results}
\end{table*}

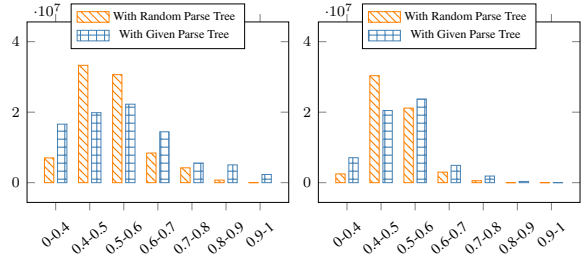
\begin{figure}[t!]
\centering
\begin{tikzpicture}[scale=0.7]
\pgfplotsset{width=6.5cm, height=5cm, compat=1.3}
\begin{axis}[
    xtick={1,2,3,4,5,6,7,8,9,10},
    ymin=0, ymax=4e7,
    xticklabels = {0-0.4, 0.4-0.5, 0.5-0.6, 0.6-0.7,0.7-0.8,0.8-0.9,0.9-1},
    xticklabel style = {font=\fontsize{8}{1}\selectfont, rotate=40},
    yticklabel style = {font=\fontsize{8}{1}\selectfont, },
    legend style={font=\fontsize{6.5}{1}\selectfont},
	y label style={at={(axis description cs:-0.12,.5)},anchor=south, xshift=2cm},
	enlargelimits=0.14,
	legend style={at={(0.5,0.85)},anchor=south,legend columns=1}, 
	label style={font=\fontsize{8}{1}\selectfont},
	ybar=2pt,
	bar width=5pt,
    ]
\addplot[orange,pattern=north west lines, pattern color=orange,  area legend] coordinates {
(1, 7031570) (2, 33294755) (3, 30655443) (4,8406454) (5,4213372) (6, 730254)(7, 14441)
};
 \addplot[skyblue, pattern=grid, pattern color=skyblue, area legend]  coordinates {
(1, 16599921) (2, 19857437) (3, 22265879) (4,14432985) (5,5574856) (6, 5049848)(7, 2324900)
};

\legend{ With Random Parse Tree\\ With Given Parse Tree\\}
\end{axis}
\end{tikzpicture}
\begin{tikzpicture}[scale=0.7]
\pgfplotsset{width=6.5cm, height=5cm, compat=1.3}
\begin{axis}[
    xtick={1,2,3,4,5,6,7,8,9,10},
    ymin=0, ymax=4e7,
    xticklabels = {0-0.4, 0.4-0.5, 0.5-0.6, 0.6-0.7,0.7-0.8,0.8-0.9,0.9-1},
    xticklabel style = {font=\fontsize{8}{1}\selectfont, rotate=40},
    yticklabel style = {font=\fontsize{8}{1}\selectfont, },
    legend style={font=\fontsize{6.5}{1}\selectfont},
	y label style={at={(axis description cs:-0.12,.5)},anchor=south, xshift=2cm},
	enlargelimits=0.14,
	legend style={at={(0.5,0.85)},anchor=south,legend columns=1}, 
	label style={font=\fontsize{8}{1}\selectfont},
	ybar=2pt,
	bar width=5pt,
    ]
\addplot[orange,pattern=north west lines, pattern color=orange,  area legend] coordinates {
(1, 2479820) (2, 30351827) (3, 21155065) (4,2987867) (5,562725) (6, 3253)(7,105)
 };
 \addplot[skyblue, pattern=grid, pattern color=skyblue, area legend]  coordinates {
(1, 7087624) (2, 20462418) (3, 23745899) (4,4913401) (5,1880041) (6, 351760)(7, 6038)
};

\legend{ With Random Parse Tree\\ With Given Parse Tree\\}
\end{axis}
\end{tikzpicture}
\vspace{-3mm}
\caption{Left: English, Right: Chinese. The x-axis indicates the value of gate $m_t$, the y-axis denotes the number of cells. }
\label{fig:rob}
\vspace{-3mm}
\end{figure}

\begin{figure}[t!]
\centering
\begin{tikzpicture}[scale=0.65]
\pgfplotsset{width=6.5cm, height=5cm, compat=1.3}
\begin{axis}[
    xtick={1,2,3,4,5},
    ymin=80, ymax=95,
    xticklabels = {$\leq$ 14, 15 - 29 ,30 - 44 , 45 - 59 , $\geq$ 60},
    xticklabel style = {font=\fontsize{8}{1}\selectfont, rotate=0, xshift=0mm},
    yticklabel style = {font=\fontsize{8}{1}\selectfont, },
    legend style={font=\fontsize{6.5}{1}\selectfont},
	y label style={at={(axis description cs:-0.12,.5)},anchor=south, xshift=2cm},
	legend style={at={(0.64,0.355)}},
	label style={font=\fontsize{9}{1}\selectfont},
    ]
    \addplot [mark=square, mark size=1.5pt, color=fruitpurple] plot coordinates {
    (1, 88.8) (2, 90.5) (3, 90.4) (4, 89.3) (5, 89.6) };
    \addplot [mark=triangle, mark size=1.5pt, color=skyblue] plot coordinates {
    (1, 89.4) (2, 90.3) (3, 90.5) (4, 90.6) (5, 89.5) };
    \addplot [mark=o, mark size=1.5pt, color=orange] plot coordinates {
    (1, 91.0) (2, 91.5) (3, 91.6) (4, 91.3) (5, 89.9) };
\legend{ BiLSTM-CRF$_{\scaleto{\text{+ BERT}}{3.5pt}}$\\ DGLSTM-CRF$_{\scaleto{\text{+ ELMO}}{3.5pt}}$\\ Syn-LSTM-CRF$_{\scaleto{\text{+ BERT}}{3.5pt}}$\\}
\end{axis}
\end{tikzpicture}
\begin{tikzpicture}[scale=0.65]
\pgfplotsset{width=6.5cm, height=5cm, compat=1.3}
\begin{axis}[
    xtick={1,2,3,4,5},
    ymin=70, ymax=90,
    xticklabels = {$\leq$ 14, 15 - 29 ,30 - 44 , 45 - 59 , $\geq$ 60},
    xticklabel style = {font=\fontsize{8}{1}\selectfont, rotate=0, xshift=0mm},
    yticklabel style = {font=\fontsize{8}{1}\selectfont, },
    legend style={font=\fontsize{6.5}{1}\selectfont},
	y label style={at={(axis description cs:-0.12,.5)},anchor=south, xshift=2cm},
	legend style={at={(0.64,0.355)}}, 
	label style={font=\fontsize{9}{1}\selectfont},
    ]
    \addplot [mark=square, mark size=1.5pt, color=fruitpurple] plot coordinates {
    (1, 83.8) (2, 81.1) (3, 82.8) (4, 81.0) (5, 79.7) };
    \addplot [mark=triangle, mark size=1.5pt, color=skyblue] plot coordinates {
    (1, 83.7) (2,81.2) (3, 82.2) (4, 80.8) (5, 79.4) };
    \addplot [mark=o, mark size=1.5pt, color=orange] plot coordinates {
    (1, 85.0) (2, 81.5) (3, 82.6) (4, 81.4) (5, 80.0) };
\legend{ BiLSTM-CRF$_{\scaleto{\text{+ BERT}}{3.5pt}}$\\ DGLSTM-CRF$_{\scaleto{\text{+ ELMO}}{3.5pt}}$\\ Syn-LSTM-CRF$_{\scaleto{\text{+ BERT}}{3.5pt}}$\\}
\end{axis}
\end{tikzpicture}
\vspace{-2mm}
\caption{Left: English, Right: Chinese. 
$x$-axis: sentence length. $y$-axis:$F_1$ score (\%). Note that DGLSTM-CRF$_{\scriptscriptstyle \text{+ ELMO}}$ have better performance compared to  DGLSTM-CRF$_{\scriptscriptstyle \text{+ BERT}}$ based on the results in the main paper.
}
\label{fig:sentlen}
\vspace{-3mm}
\end{figure}
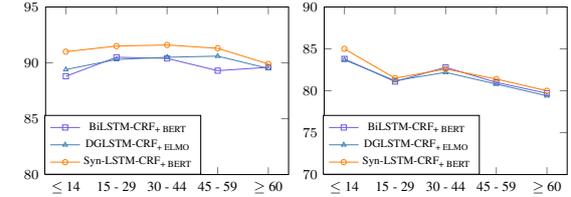

\section{Performance of dependency parser}
Table \ref{tab:dep_result} presents the performance of dependency parser.

\section{More data statistics}
Table \ref{tab:statslength} shows the statistics of the number of entities with respect to entity length for OntoNotes 5.0 English and Chinese, SemEval 2010 Task 1 Catalan and Spanish datasets.

\begin{figure*}
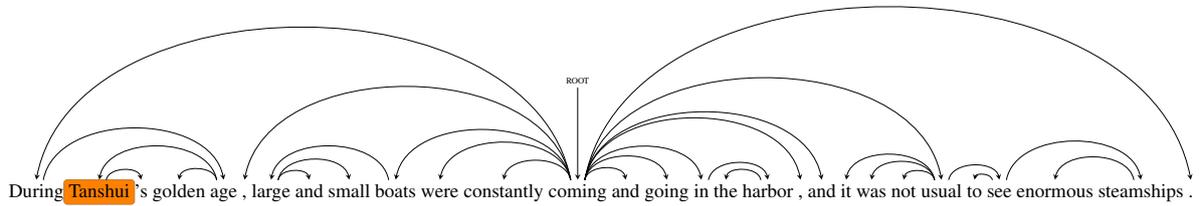

    \centering
    \adjustbox{max width=1\linewidth}{
    \begin{dependency}[theme = simple]
        \begin{deptext}[column sep=-0.2em]
        During  \& Tanshui \& 's \& golden \& age \& , \& large \& and \& small \& boats \& were \& constantly \& coming \& and \& going \& in \& the \& harbor \& , \& and \& it \& was \& not \& usual \& to \& see \& enormous \& steamships \& . \\
        \end{deptext}
        \depedge{13}{1}{}
        \depedge{5}{2}{}
        \depedge{2}{3}{}
        \depedge{5}{4}{}
        \depedge{1}{5}{}
        \depedge{13}{6}{}
        \depedge{10}{7}{}
        \depedge{7}{8}{}
        \depedge{7}{9}{}
        \depedge{13}{10}{}
        \depedge{13}{11}{}
        \depedge{13}{12}{}
        \deproot{13}{ROOT}
        \depedge{13}{14}{}
        \depedge{13}{15}{}
        \depedge{13}{16}{}
        \depedge{18}{17}{}
        \depedge{16}{18}{}
        \depedge{13}{19}{}
        \depedge{13}{20}{}
        \depedge{24}{21}{}
        \depedge{24}{22}{}
        \depedge{24}{23}{}
        \depedge{13}{24}{}
        \depedge{26}{25}{}
        \depedge{24}{26}{}
        \depedge{28}{27}{}
        \depedge{26}{28}{}
        \depedge{13}{29}{}
        \wordgroup[group style={fill=orange, draw=brown, inner sep=0ex}]{1}{2}{2}{a0}
    \end{dependency}
    }
    \caption{An example of dependency tree. The mentioned entity is highlighted in orange, and the entity type is GPE.}
    \label{fig:dependency}
\end{figure*}

\section{More Robustness Analysis}
Figure \ref{fig:rob} shows the comparisons of the models of using random trees and given trees on OntoNotes 5.0 English and Chinese datasets.

\section{Effect of Sentence Length}
We compare the performance of our Syn-LSTM-CRF$_{\scriptscriptstyle \text{+ BERT}}$ with BiLSTM-CRF$_{\scriptscriptstyle \text{+ BERT}}$ and DGLSTM-CRF$_{\scriptscriptstyle \text{+ ELMO}}$ models with respect to sentence length, and the results are shown in Figure \ref{fig:sentlen}.

\section{Case Study}
We further show an example to visualize the propagation of non-local information (Figure \ref{fig:dependency}). The example is selected from OntoNotes 5.0 English dataset. Even though the DGLSTM-CRF \cite{Jie2019DependencyGuidedLF} model is able to recognize "Tianshui" as a named entity, it predicts a wrong entity type as PERSON while the true type is GPE. If only looking at the first half of the sentence, it is possible to predict "Tianshui" as PERSON because of the local information "age". However, the second half of the sentence confirms that the entity type of "Tianshui" is GPE. With the non-local information from the graph-encoded representation, our Syn-LSTM-CRF successfully predicts the right entity type.

\end{document}